\documentclass[10pt,twocolumn,letterpaper]{article}

\usepackage{iccv}

\usepackage{times}
\usepackage{epsfig}
\usepackage{graphicx}
\usepackage{amsmath}
\usepackage{amssymb}
\usepackage{pifont}
\usepackage{color}
\usepackage{multirow}
\usepackage{overpic}
\usepackage{arydshln}
\usepackage{makecell}
\usepackage{times}
\usepackage[T1]{fontenc}
\usepackage{hhline}
\usepackage{xcolor}
\usepackage{pifont}
\usepackage{enumitem}
\usepackage{colortbl}
\usepackage{verbatim}
\usepackage{bm}
\usepackage{wrapfig}
\usepackage{booktabs}
\usepackage{eso-pic}
\usepackage{placeins}
\usepackage[nopar]{lipsum}
\usepackage{capt-of}
\usepackage[accsupp]{axessibility}

\def\eg{\emph{e.g}\onedot} 
\def\ie{\emph{i.e}\onedot}

\def\etal{\emph{et al}\onedot}

\definecolor{darkpastelgreen}{rgb}{0.01, 0.75, 0.24}
\definecolor{darkpink}{rgb}{0.91, 0.33, 0.5}
\definecolor{mygray}{gray}{.94}

\definecolor{linkcolor}{RGB}{255,0,0}
\definecolor{urlcolor}{RGB}{255,105,180}
\definecolor{citecolor}{RGB}{0, 80, 200}
\definecolor{citecolor1}{RGB}{0,153,255}

\newcommand\blfootnote[1]{%
  \begingroup
  \renewcommand\thefootnote{}\footnote{#1}%
  \addtocounter{footnote}{-1}%
  \endgroup
}

\usepackage[pagebackref=true,breaklinks=true,letterpaper=true,colorlinks,bookmarks=false]{hyperref}

\iccvfinalcopy

\begin{document}

\title{Grounding 3D Object Affordance from 2D Interactions in Images}

\author{Yuhang Yang$^{1}$, Wei Zhai$^{1,*}$, Hongchen Luo$^{1}$, Yang Cao$^{1,2}$, Jiebo Luo$^{3}$, Zheng-Jun Zha$^{1}$\\
{$^{1}$~University of Science and Technology of China} \qquad
{$^{3}$~University of Rochester}\\
{$^{2}$~Institute of Artificial Intelligence, Hefei Comprehensive National Science Center}\\
\small{\texttt{\{yyuhang@mail., wzhai056@mail., lhc12@mail., forrest@\}ustc.edu.cn}} \\
\small{\texttt{jluo@cs.rochester.edu}} \qquad
\small{\texttt{zhazj@ustc.edu.cn}}
}

\maketitle
\blfootnote{$*$Corresponding Author.}

\begin{abstract}
Grounding 3D object affordance seeks to locate objects' ``action possibilities'' regions in the 3D space, which serves as a link between perception and operation for embodied agents. Existing studies primarily focus on connecting visual affordances with geometry structures, \eg, relying on annotations to declare interactive regions of interest on the object and establishing a mapping between the regions and affordances. However, the essence of learning object affordance is to understand how to use it, and the manner that detaches interactions is limited in generalization. Normally, humans possess the ability to perceive object affordances in the physical world through demonstration images or videos. Motivated by this, we introduce a novel task setting: grounding 3D object affordance from 2D interactions in images, which faces the challenge of anticipating affordance through interactions of different sources. To address this problem, we devise a novel Interaction-driven 3D Affordance Grounding Network (IAG), which aligns the region feature of objects from different sources and models the interactive contexts for 3D object affordance grounding. Besides, we collect a Point-Image Affordance Dataset (PIAD) to support the proposed task. Comprehensive experiments on PIAD demonstrate the reliability of the proposed task and the superiority of our method. The code, model and data are available at the \href{https://yyvhang.github.io/publications/IAG/index.html}{project page.} 
\end{abstract}

\section{Introduction}
The term ``affordance'' is described as ``opportunities of interaction'' by J. Gibson \cite{gibson2014ecological}. Grounding 3D object affordance aims to comprehend the interactive regions of objects in 3D space, which is not only simply to predict which interaction an object affords, but also to identify specific points on the object that could support the interaction. It constitutes a link between perception and operation for embodied agents, which has the potential to serve numerous practical applications, \eg action prediction \cite{koppula2013learning, vu2014predicting}, robot manipulation \cite{mandikal2021learning, moldovan2012learning, osiurak2016tool}, imitation learning \cite{hussein2017imitation, osa2018algorithmic}, and augmented/virtual reality \cite{cheng2013affordances, dalgarno2010learning}.

\begin{figure}[t]
	\centering
	\small
		\begin{overpic}[width=0.92\linewidth]{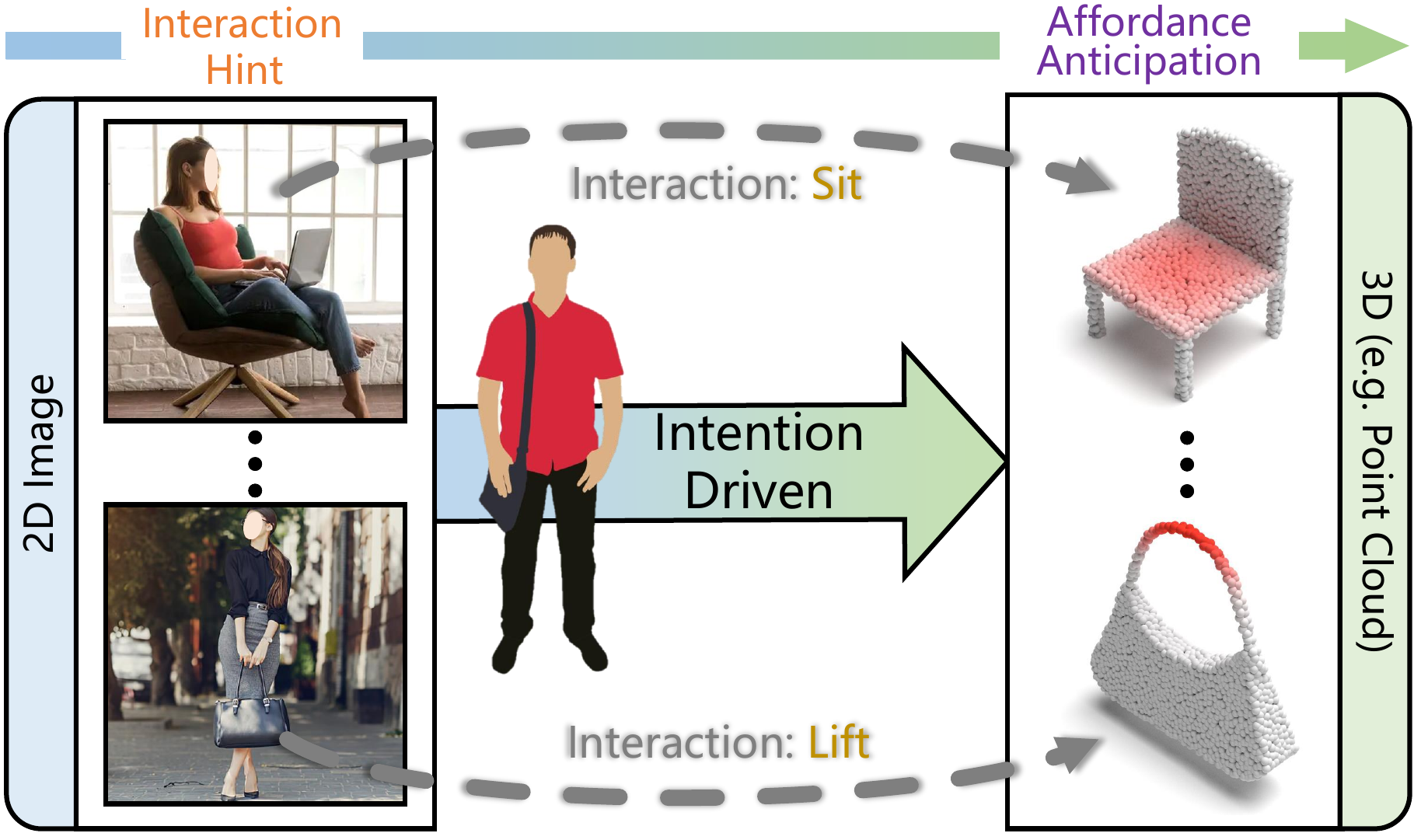}
		    
	\end{overpic}
	\caption{\textbf{Grounding Affordance from Interactions.} We propose to ground 3D object affordance through 2D interactions. Given an object point cloud with an interactive image, grounding the corresponding affordance on the 3D object.}
 \label{fig1}
\end{figure}

\begin{figure*}[t]
	\centering
	\small
		\begin{overpic}[width=0.89\linewidth]{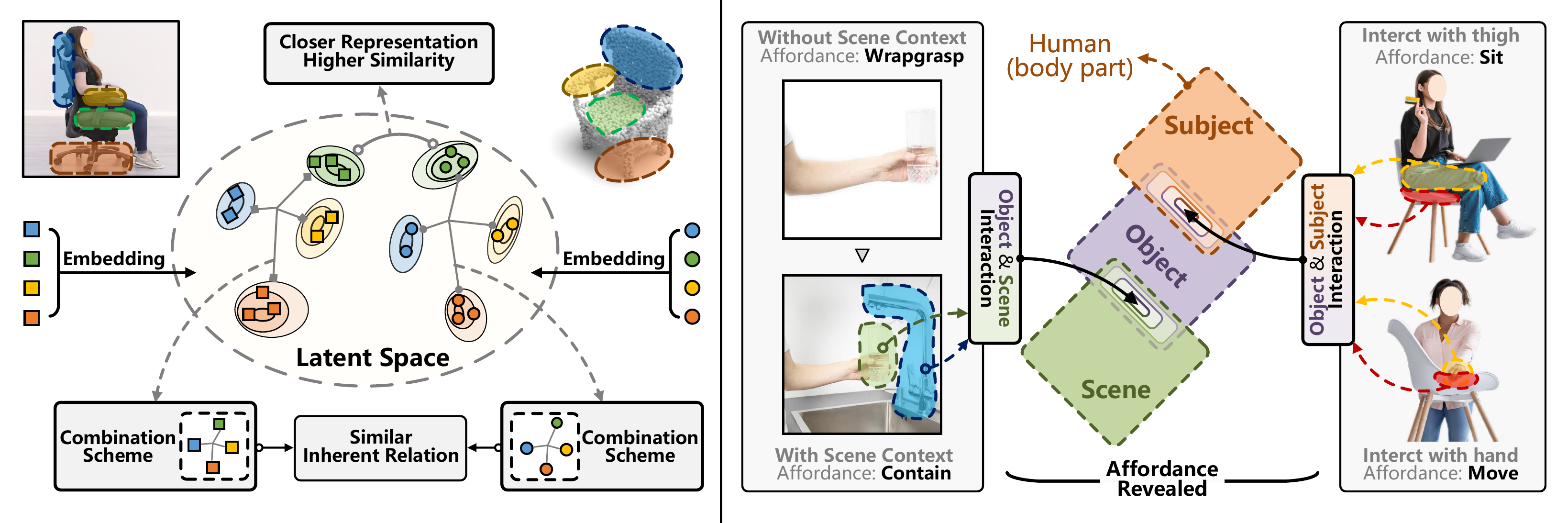}
		\put(23,-0.3){\textbf{(a)}}
		\put(73,-0.3){\textbf{(b)}}
	\end{overpic}
	\caption{\textbf{Motivation.} 
        \textbf{(a)} Clues to correlate regions of objects from different sources. Objects with the same category possess a similar combination scheme to meet certain affordance properties, it is a similar inherent relation among different instances.  And some structures hint at these properties exhibit closer representations with higher similarity in the latent space. \textbf{(b)} Object affordance may be affected by dynamic factors, such as the position of the object itself in the scene, other objects in the scene, and the body part of the interacting subject. These factors can be decomposed into object-subject and object-scene interaction contexts.
	}
 \label{Fig:motivation}
\end{figure*}

\par So far, the paradigm of perceiving 3D object affordance has several branches. One of them involves establishing an explicit mapping between affordance categories and geometry structures \cite{deng20213d, kim2014semantic, myers2015affordance, xu2022partafford}, based on visual appearance. However, affordance is dynamic and multiple, these geometric-specific manners have limited generalization for unseen structures. Besides, locking the geometry with a specific affordance category may lead to the anticipated region being inconsistent with its affordance when objects possess multiple similar geometrics, resulting in affordance regional confusion. Another paradigm is based on reinforcement learning, which puts the agent in 3D synthetic scenarios to interact with several objects actively, taking the reward mechanism to optimize the whole process \cite{nagarajan2020learning}. While this type of approach transitions agents from passive recognition to active reasoning, it needs repeated attempts in a huge search space when meeting a novel structure, and is therefore time-consuming. 
\par These limitations motivate us to explore an affordance-compatible learning paradigm. Typically, humans can infer object affordance in the 3D physical world by watching images or videos that demonstrate the interactions. Some studies in cognitive science \cite{merleau1976phenomenologie, shilling2004body}  point out the existence of ``body image'' in human cognition, which claims that humans have a perceptual experience of the objects they see, thus facilitating their ability to organize the perception \cite{zhai2022exploring} and operate novel objects. Hence, human-object interaction conveys the perception that object structure could perform certain affordance, which is a crucial clue to reason object affordance. In light of this, we present a novel task setting: grounding 3D object affordance from 2D interactions in images, which is shown in Fig. \ref{fig1}.

\par This challenging task includes several essential issues that should be properly addressed. \textbf{1) Alignment ambiguity}. To ground 3D object affordance from 2D source interactions, the premise is to correspond the regions of the object in different sources. The object in 2D demonstrations and the 3D object we face are usually derived from different physical instances in different locations and scales. This discrepancy may lead to confusion when matching the affordance regions, causing alignment ambiguity. While objects are commonly designed to satisfy certain needs of human beings, so the same category generally follows a similar combination scheme of object components to meet certain affordances, and these affordances are hinted at by some structures (Fig. \ref{Fig:motivation} (a)). These invariant properties are across instances and could be utilized to correlate object regions from different sources. \textbf{2) Affordance ambiguity}. Affordance has properties of dynamic and  multiplicity, which means the object affordance may change according to the situation, the same part of an object could afford multiple interactions, as shown in Fig. \ref{Fig:motivation} (b), ``Chair'' affords ``Sit'' or ``Move'' depends on the human actions, ``Mug'' affords ``Wrapgrasp'' or ``Contain'' according to the scene context, these properties may make ambiguity when extracting affordance. However, these dynamic factors for affordance extraction can be decomposed into the interaction between subject-object and object-scene. Modeling these interactions is possible to extract explicit affordance.

\par To address these issues, we propose the \textbf{I}nteraction-driven 3D \textbf{A}ffordance \textbf{G}rounding Network (\textbf{IAG}) to align object region features from different sources and model interaction contexts to reveal affordance. In detail, it contains two sub-modules, one is \textbf{J}oint \textbf{R}egion \textbf{A}lignment Module (\textbf{JRA}), which is devised to eliminate alignment ambiguity. It takes the relative difference in dense cross-similarity to refine analogous shape regions and employs learnable layers to map the invariant combination scheme, taking them to match local regions of objects. For affordance ambiguity, the other one \textbf{A}ffordance \textbf{R}evealed \textbf{M}odule (\textbf{ARM}) takes the object representation as a shared factor and jointly models its interaction contexts with the affordance-related factors to reveal explicit affordance. Moreover, we collect \textbf{P}oint-\textbf{I}mage \textbf{A}ffordance \textbf{D}ataset (\textbf{PIAD}) that contains plenty of paired image-point cloud affordance data and make a benchmark to support the model training and evaluation on the proposed setting.

\par The contributions are summarized as follows: \textbf{1)} We introduce grounding 3D object affordance through the 2D interactions, which facilitates the generalization to 3D object affordance perception. \textbf{2)} We propose the IAG framework, which aligns the region feature of objects from different sources and jointly models affordance-related interactions to locate the 3D object affordance. \textbf{3)} We collect a dataset named PIAD to support the proposed task setting, which contains paired image-point cloud affordance data. Besides, we establish a benchmark on PIAD, and the experiments on PIAD exhibit the reliability of the method and the setting.
\section{Related Works}
\label{sec:relation work}
\subsection{Affordance Learning}
\par At present, the affordance area has made achievements in multiple tasks (Tab. \ref{tab:related}). Some works are devoted to detecting the affordance region from 2D sources \ie images and videos \cite{chuang2018learning, do2018affordancenet, Li2023G2L, luo2021one, roy2016multi, thermos2020deep, zhao2020object}, and there are also some studies accomplish this task with the assistance of natural language \cite{lu2022phrase, mi2020intention, mi2019object}. They seek to detect or segment objects that afford the action in 2D data, but cannot perceive the specific parts corresponding to the affordance. Thus, another type of method brings a leap to locate the affordance region of objects from 2D sources \cite{fang2018demo2vec, luo2022learning, nagarajan2019grounded, luo2022grounded}. However, the affordance knowledge derived from 2D sources is hard to extrapolate specific interactive locations of objects in the 3D environment. With several 3D object datasets proposed \cite{Geng_2023_CVPR, Liu_2022_CVPR, Mo_2019_CVPR}, some researchers explore grounding object affordance from 3D data \cite{deng20213d, mo2022o2o, xu2022partafford, nagarajan2020learning}. Such methods directly map the connection between semantic affordances and 3D structures, detach the real interaction, and may deter the generalization ability: structures that do not map to a specific affordance are usually hard to generalize through this type of method. T. Nagarajan \etal \cite{nagarajan2020learning} give a fresh mind, taking the reinforcement learning to make agents actively interact with the 3D synthetic scenarios, while it requires repeated attempts by the agents in a given search space, and is time-consuming. In robotics, affordance is utilized to provide priors for object manipulation and achieve considerable results, especially the articulated 3D objects \cite{mo2021where2act,wang2022adaafford, zhao2022dualafford}. Several methods utilize the 2.5D data \eg RGB-D image to understand the object affordance and serve the manipulation task \cite{koppula2013learning, koppula2014physically, koppula2015anticipating, nguyen2016detecting}, for these methods, the image and depth information are corresponding in spatial and need to be collected in the same scene. In contrast, our task focus on grounding 3D object affordance from 2D interactions, in which the 2D interactions provide direct clues to excavate object affordance efficiently and make the affordance could generalize to some unseen structures, and the 2D-3D data is collected from different sources, freeing the constraints on spatial correspondence. 

\subsection{Image-Point Cloud Cross-Modal Learning}
\par The combination of point cloud and image data can capture both semantic and geometric information, enabling cross-modal learning among them has great potential application value in scenarios like autopilot \cite{cui2021deep}. For this type of work, aligning features is the premise for completing downstream tasks \eg detection, segmentation, and registration. It is aimed at establishing correspondences between instances from different modalities \cite{baltruvsaitis2018multimodal, Li2023U, yang2023implicit}, either spatial or semantic. To achieve this, many methods use camera intrinsics to correspond spatial position of pixels and points, then align per pixel-point feature or fuse the raw data \cite{krispel2020fuseseg, tan2021mbdf, vora2020pointpainting,xu2022fusionrcnn, zhao2021lif, zhuang2021perception}. Some works utilize depth information to project image features into 3D space and then fuse them with point-wise features \cite{jaritz2019multi, mithun2020rgb2lidar, yin2021multimodal, zhang2020fusion, zhao2021similarity}. The above methods rely on the spatial prior information to align features, while in our task, images and point clouds are collected from distinct physical instances, there are no priors on camera pose or intrinsics, also no corresponding depth information of the image. Therefore, we align the object region features that are derived from different sources in the feature space with the assistance of the correlation invariance between affordances and appearance structures.

\begin{table}[t]
\centering
\small
  \renewcommand{\arraystretch}{1.}
  \renewcommand{\tabcolsep}{4.pt}
\caption{\textbf{Affordance Learning.} Various works for several tasks in the affordance community.
}
\label{tab:related}
\vspace{5pt}
\begin{tabular}{c|c|c|c}
\toprule
\textbf{Work} & \textbf{Input}             & \textbf{Output}             & \textbf{Task}         \\ \midrule
 \cite{do2018affordancenet, thermos2020deep, zhao2020object}            & Image/Video          & 2D Mask         & Detection    \\ 
\cite{lu2022phrase, mi2020intention, mi2019object}             & Image,Text          & 2D Mask         & Detection    \\ 
\cite{koppula2013learning, koppula2015anticipating, nguyen2016detecting}             & RGBD             & Heatmap \& Action        &  Manipulation\\
\cite{mo2021where2act,wang2022adaafford, zhao2022dualafford}    & Point Cloud & Heatmap \& Action & Manipulation    \\
\cite{fang2018demo2vec, luo2022learning, nagarajan2019grounded}             & Image/Video       & 2D Heatmap         & Grounding    \\
\cite{deng20213d, mo2022o2o, xu2022partafford}     & Point Cloud       & 3D Heatmap & Grounding    \\
\bottomrule
\end{tabular}
\end{table}

\section{Method}
\label{sec:method}
\subsection{Overview}
\label{Section:3.1}
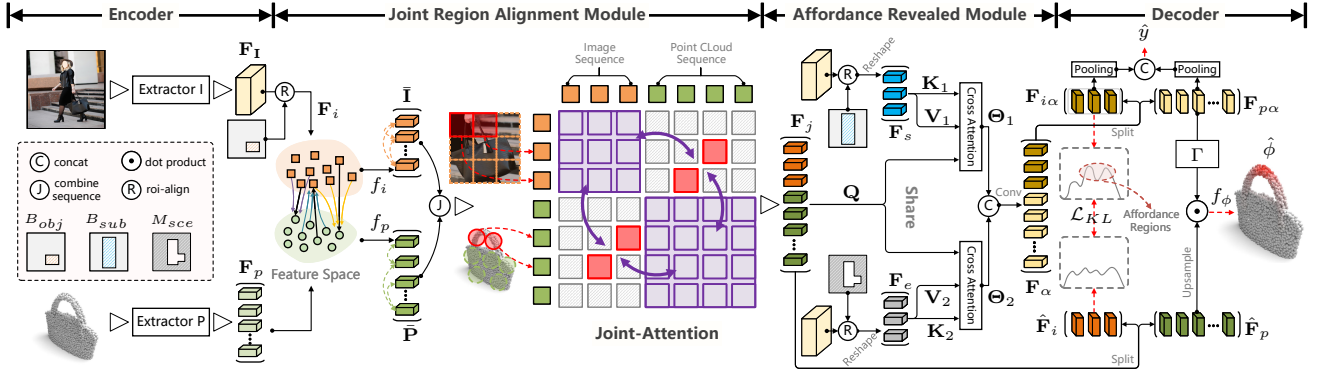
\begin{figure*}[t]
	\centering
        \scriptsize
	\begin{overpic}[width=1.\linewidth]{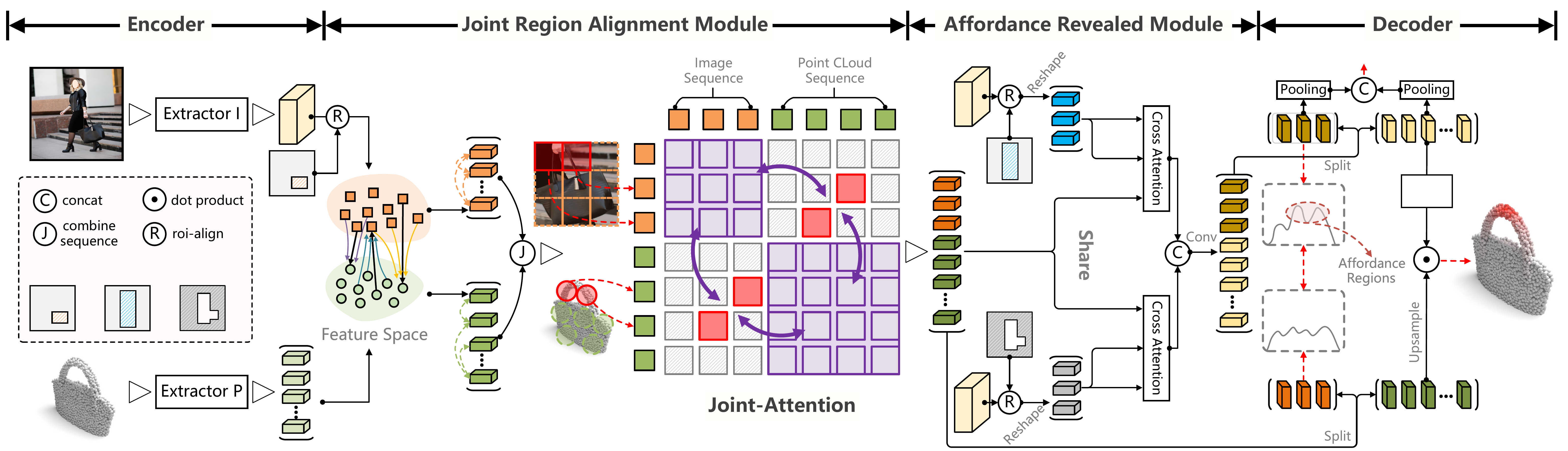}
        \put(28,11){$f_p$}
        \put(28,13.8){$f_i$}
        \put(30.5,2.3){$\bar{\mathbf{P}}$}
        \put(30.5,20.5){$\bar{\mathbf{I}}$}
        \put(18,24.1){$\mathbf{F_{I}}$}
        \put(24,19.5){$\mathbf{F}_i$}
        \put(18,7.5){$\mathbf{F}_p$}
        \put(1.7,11.3){\tiny$B_{obj}$}
        \put(6.3,11.3){\tiny$B_{sub}$}
        \put(11.3,11.2){\tiny$M_{sce}$}
        \put(60,18.6){\scriptsize$\mathbf{F}_{j}$}
        \put(67.5,17.8){\scriptsize$\mathbf{F}_{s}$}
        \put(67.5,6.5){\scriptsize$\mathbf{F}_{e}$}
        \put(77.9,6){\scriptsize$\mathbf{F}_{\alpha}$}
        \put(77.9,20.5){\scriptsize$\mathbf{F}_{i\alpha}$}
        \put(94.5,20){\scriptsize$\mathbf{F}_{p\alpha}$}
        \put(78.7,3){\scriptsize$\hat{\mathbf{F}}_{i}$}
        \put(94.6,2.8){\scriptsize$\hat{\mathbf{F}}_{p}$}  
        \put(64,13.2){\scriptsize $\mathbf{Q}$}
        \put(70,21.5){\scriptsize $\mathbf{K}_1$}
        \put(70.2,19){\scriptsize $\mathbf{V}_1$}
        \put(70.4,2.6){\scriptsize $\mathbf{K}_2$}
        \put(70.2,5.2){\scriptsize $\mathbf{V}_2$}
        \put(75,19){\scriptsize $\mathbf{\Theta}_{1}$}
        \put(75,5.2){\scriptsize $\mathbf{\Theta}_{2}$}
        \put(90.5,16){\scriptsize$\Gamma$}
        \put(92,13){\scriptsize$f_{\phi}$}
        \put(86.5,25.5){\scriptsize $\hat{y}$}
        \put(96,16.5){\scriptsize $\hat{\phi}$}
        \put(81.3,11.5){\scriptsize $\mathbf{\mathcal{L}}_{KL}$}
	\end{overpic}
	\caption{\textbf{Method.} Overview of Interaction-driven 3D Affordance Grounding Network (IAG), it firstly extracts localized features $\mathbf{F}_{i}, \mathbf{F}_{p}$ respectively, then takes JRA (Sec. \ref{Section:3.3}) to align them and get the joint feature $\mathbf{F}_{j}$. Next, ARM (Sec. \ref{Section: 3.4}) utilizes $\mathbf{F}_{j}$ to reveal affordance $\mathbf{F}_{\alpha}$ with $\mathbf{F}_s, \mathbf{F}_e$ by cross-attention. Eventually, $\mathbf{F}_{j}$ and $\mathbf{F}_{\alpha}$ are sent to the decoder (Sec. \ref{section3.5}) to obtain the final results $\hat{\phi}$ and $\hat{y}$.}
 \label{fig:method}
\end{figure*}

Our goal is to anticipate the affordance regions on the point cloud that correspond to the interaction in the image. Given a sample $\{P,I,\mathcal{B}, y\}$, where $P$ is a point cloud with the coordinates $P_{c} \in \mathbb{R}^{N \times 3}$ and the affordance annotation $P_{label} \in \mathbb{R}^{N \times 1}$, $I \in \mathbb{R}^{3 \times H \times W}$ is an RGB image, $\mathcal{B}=\{B_{sub},B_{obj}\}$ denotes the bounding box of the subject and object in $I$, and $y$ is the affordance category label. The \textbf{IAG} (Fig. \ref{fig:method}) capture localized features $\mathbf{F}_{\mathbf{I}} \in \mathbb{R}^{C \times H^{'} \times W^{'}}$ and $\mathbf{F}_p \in \mathbb{R}^{C \times N_{p}}$ of the image and point cloud by two feature extractors ResNet \cite{he2016deep} and PointNet++ \cite{qi2017pointnet++}. Then, utilizing $B_{obj}/B_{sub}$ to locate the object/subject region in $\mathbf{F}_{\mathbf{I}}$, outside the $B_{obj}$ and $B_{sub}$ is the scene mask $M_{sce}$, use ROI-Align \cite{he2017mask} to get the object, subject, and scene features $\mathbf{F}_{i}, \mathbf{F}_{s}, \mathbf{F}_{e} \in \mathbb{R}^{C \times H_{1} \times W_{1}}$, and reshape them to $\mathbb{R}^{C \times N_{i}}$ ($N_i = H_1 \times W_1$). Next, the JRA module takes $\mathbf{F}_i, \mathbf{F}_p$ as input and jointly mines the correspondence to align them, obtaining the joint feature $\mathbf{F}_j$ (Sec. \ref{Section:3.3}). Following, the ARM module takes $\mathbf{F}_j, \mathbf{F}_s, \mathbf{F}_e$ to reveal the affordance $\mathbf{F}_{\alpha}$ through cross-attention mechanism (Sec. \ref{Section: 3.4}). Ultimately, $\mathbf{F}_{\alpha}$ and $\mathbf{F}_j$ are sent to the decoder to compute the affordance logits $\hat{y}$ and the final 3D object affordance $\hat{\phi}$ (Sec. \ref{section3.5}). The process is expressed as $\hat{\phi}, \hat{y} = f_{\theta}(P_{c},I,\mathcal{B};\theta)$, where $f_{\theta}$ denotes the network and $\theta$ is the parameter.

\subsection{Joint Region Alignment Module}
\label{Section:3.3}
The JRA module calculates the high-dimensional dense similarity in feature space to approximate analogous shape regions that contain relatively higher similarity \cite{baltruvsaitis2018multimodal}. Meanwhile, to map the inherent combination scheme of object components, the JRA models the inherent relations between modality-specific regions and takes it to drive the alignment of other regions by a transformer-based joint-attention $f_\delta$. With the network trained to capture the correspondences among regions from different sources, the alignment is performed implicitly during the optimization.
\par Initially, $\mathbf{F}_p$ and $\mathbf{F}_i$ are projected into a feature space by shared convolution layers $f_\sigma$, obtaining the features $\mathbf{P} \in \mathbb{R}^{C \times N_{p}}, \mathbf{I} \in \mathbb{R}^{C \times N_{i}}$. Then, to correlate the analogous shape regions, the dense cross-similarity is calculated between each region of $\mathbf{P}$ and $\mathbf{I}$, formulated as:
\begin{equation}
\label{Equ:1}
    \varphi_{i,j}=\frac{e^{(\mathbf{P}_{i},\mathbf{I}_{j})}}{\sum^{N_p}_{i=1}\sum^{N_i}_{j=1}e^{(\mathbf{P}_i,\mathbf{I}_{j})}}, \varphi \in \mathbb{R}^{N_p \times N_i},
\end{equation}
where $\varphi_{i,j}$ denotes the cross-similarity between the i-th region of $\mathbf{P}$ and the j-th region of $\mathbf{I}$. Taking the relative difference in $\varphi$ to refine and correlate analogous regions in $\mathbf{P}$ and $\mathbf{I}$, next, applying self-attention layers $f_i, f_p$ to model the intra-structural inherent relation of objects in respective modalities, the process is expressed as:
\begin{equation}
    \bar{\mathbf{P}} = f_p(\mathbf{I} \cdot \varphi^{T}), \bar{\mathbf{I}} = f_i(\mathbf{P} \cdot \varphi),
\end{equation}
where $\bar{\mathbf{P}} \in \mathbb{R}^{C\times N_p}, \bar{\mathbf{I}} \in \mathbb{R}^{C\times N_i}$. Following, we perform a joint-attention on features with structural relevance to map the similar inherent combination scheme and drive the alignment of the remaining regions. The process is formulated as: $\mathbf{F}_{j} = f_\delta[\bar{\mathbf{P}},\bar{\mathbf{I}}]$, where $\mathbf{F}_j \in \mathbb{R}^{C \times (N_{p}+N_{i})}$ and $[\cdot]$ denotes joining the image feature sequence and point cloud feature sequence into a whole one, and $\mathbf{F}_{j}$ denotes the joint representation. The $\mathbf{F}_{j}$ is sent to the affordance revealed module to excavate interaction contexts with $\mathbf{F}_{s}, \mathbf{F}_{e}$. And the affordance knowledge revealed by the interaction contexts could mutually optimize the alignment process during training (see Sec. \ref{section3.5}).

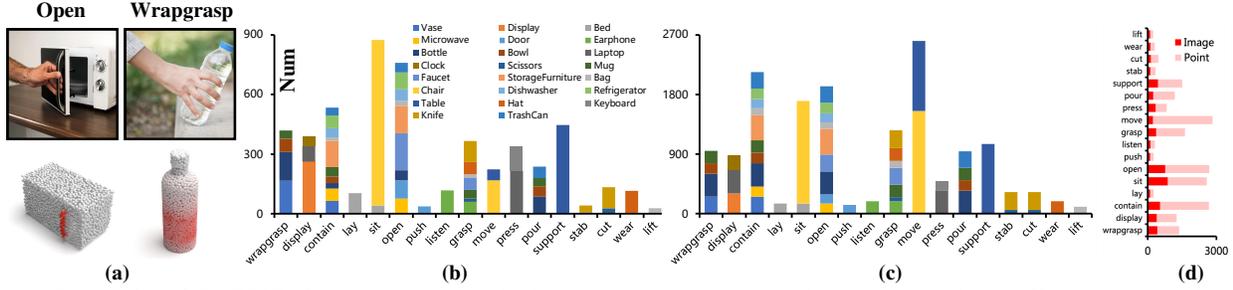
\begin{figure*}[t]
	\centering
	\footnotesize
        \begin{overpic}[width=0.95\linewidth]{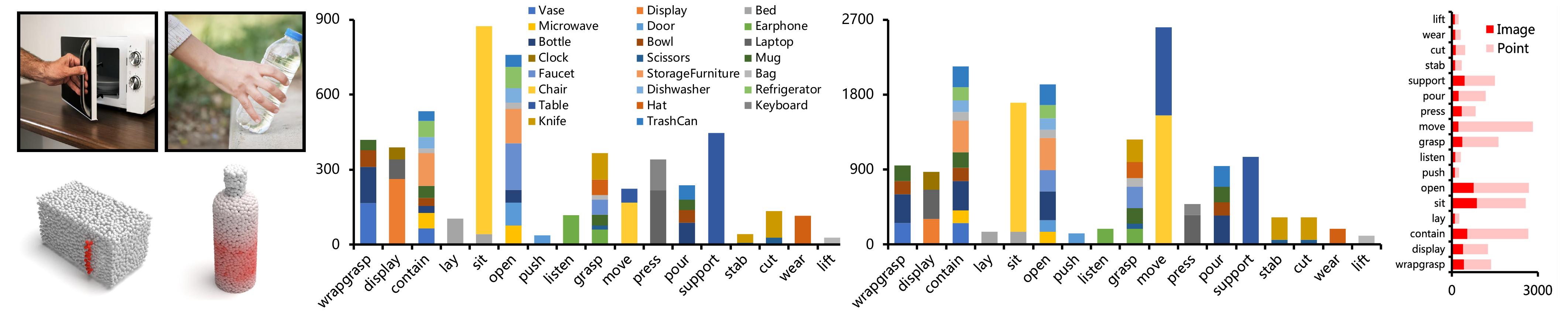}
        \put(23, 15){{\textbf{\rotatebox{90}{Num}}}}
        \put(9, 0){{\textbf{(a)}}}
        \put(36, 0){{\textbf{(b)}}}
        \put(71, 0){{\textbf{(c)}}}
        \put(95, 0){{\textbf{(d)}}}
        \put(3.5, 21){{\textbf{Open}}}
        \put(11, 21){{\textbf{Wrapgrasp}}}
	\end{overpic}
	\caption{\textbf{Properties of the PIAD dataset.} \textbf{(a)} Data pairs in the PIAD, the red region in point clouds is the affordance annotation. \textbf{(b)} Distribution of the image data. The horizontal axis represents the category of affordance, the vertical axis represents quantity, and different colors represent different objects. \textbf{(c)} Distribution of the point cloud data. \textbf{(d)} The ratio of images and point clouds in each affordance class. It shows that images and point clouds are not fixed one-to-one pairing, they can form multiple pairs. Please refer to supp. for more details.}
 \label{Fig:dataset}
\end{figure*}

\subsection{Affordance Revealed Module}
\label{Section: 3.4}
\par In general, the dynamic factors related to affordance are reflected in the interaction, affecting the extraction of affordance representation, and the interaction is mostly considered to exist between humans and objects. However, the interaction between objects and scenes (including other objects) may also affect affordance \cite{mo2022o2o}. To consider both of these factors, the ARM module utilizes the cross-attention technique to extract these interaction contexts respectively and jointly embed them to reveal explicit object affordance.
\par Specifically, $\mathbf{F}_{j}$ is projected to form the shared query $\mathbf{Q} = \mathbf{F}_{j}\mathbf{W}_{1}$, $\mathbf{F}_{s}$ and $\mathbf{F}_{e}$ are projected to form different keys and values $\mathbf{K}_{1/2} = \mathbf{F}_{s}\mathbf{W}_{2/3}, \mathbf{V}_{1/2} = \mathbf{F}_{e}\mathbf{W}_{4/5}$, where $\mathbf{W}_{1 \sim 5}$ are projection weights. Then cross-attention aggregates them to excavate interaction contexts, expressed as:
\begin{equation}
    \mathbf{\Theta}_{1/2} = softmax(\mathbf{Q}^{T}\cdot\mathbf{K}_{1/2}/\sqrt{d})\cdot\mathbf{V}^{T}_{1/2},
\end{equation}
where $\mathbf{Q} \in \mathbb{R}^{d \times (N_p+N_i)}, \mathbf{K}_{1/2}, \mathbf{V}_{1/2} \in \mathbb{R}^{d \times N_i}$, $d$ is the dimension of projection, $\mathbf{\Theta}_1, \mathbf{\Theta}_2$ indicate the excavated interaction contexts. Then, jointly embedding $\mathbf{\Theta}_1$ and $\mathbf{\Theta}_2$ to reveal affordance, expressed as:
\begin{equation}
\label{equ:affordance}
    \mathbf{F}_{\alpha} = f_{\xi}(\mathbf{\Theta}_{1},\mathbf{\Theta}_{2}), \mathbf{F}_{\alpha} \in \mathbb{R}^{C \times (N_p+N_i)},
\end{equation}
where $\mathbf{F}_{\alpha}$ is the joint affordance representation, it is split in the decoder to compute final results, $f_{\xi}$ denotes the concatenation, followed by a convolution layer to fuse the feature.

\subsection{Decoder and Loss Function}
\label{section3.5}
The $\mathbf{F}_j$ is split to $\hat{\mathbf{F}}_{p} \in \mathbb{R}^{C \times N_{p}}$ and $\hat{\mathbf{F}}_{i} \in \mathbb{R}^{C \times N_{i}}$, $\mathbf{F}_{\alpha}$ is split to $\mathbf{F}_{p\alpha} \in \mathbb{R}^{C \times N_{p}}$ and $\mathbf{F}_{i\alpha} \in \mathbb{R}^{C \times N_{i}}$ in the decoder. Pooling $\mathbf{F}_{p\alpha}$ and $\mathbf{F}_{i\alpha}$ respectively and concatnate them to compute the affordance logits $\hat{y}$. $\hat{\mathbf{F}}_{p}$ is upsampled to $\mathbb{R}^{C \times N}$ by Feature Propagation Layers (FP) \cite{qi2017pointnet++}, and the 3D object affordance $\hat{\phi}$ is computed as:
\begin{equation}
    \hat{\phi} = f_{\phi}(\mathbf{FP}(\hat{\mathbf{F}}_{p}) \odot \Gamma(\mathbf{F}_{p\alpha})), \hat{\phi} \in\mathbb{R}^{N \times 1},
    \label{equ:out}
\end{equation}
where $f_{\phi}$ is an output head, $\mathbf{FP}$ is the upsample layer, $\Gamma$ denotes pooling $\mathbf{F}_{p\alpha}$ and expand it to the shape of $\mathbb{R}^{C \times N}$.
\par The total loss comprises three parts: $\mathbf{\mathcal{L}}_{CE}$, $ \mathbf{\mathcal{L}}_{KL}$, $\mathbf{\mathcal{L}}_{HM}$. Where $\mathbf{\mathcal{L}}_{CE}$ computes the cross-entropy loss between $y$ and $\hat{y}$, it supervises the extraction of $\mathbf{F}_{\alpha}$ and implicitly optimizes the alignment process. To enable the network focus on the alignment of affordance regions, we apply the KL Divergence (KLD) \cite{bylinskii2018different} to constrain the distribution between $\mathbf{F}_{i\alpha}$ and $\hat{\mathbf{F}}_i$, formulated as $\mathbf{\mathcal{L}}_{KL} = (\mathbf{F}_{i\alpha} || \hat{\mathbf{F}}_i)$. The reason is that $\mathbf{F}_{i\alpha}$ exhibits the affordance distribution of each object region in the image, and the affordance-related regions keep more significant features. Constraining the feature distribution of $\hat{\mathbf{F}}_i$ to enhance the affordance region features in $\hat{\mathbf{F}}_i$, and with the region correlations established by the alignment process, $\hat{\mathbf{F}}_p$ also tends to exhibit this property, similar to distillation \cite{ jing2021amalgamating, yang2022factorizing}. Which makes the alignment and affordance extraction optimize mutually. $\mathbf{\mathcal{L}}_{HM}$ is a focal loss \cite{lin2017focal} combined with a dice loss \cite{milletari2016v}, it is calculated by $\hat{\phi}$ and $P_{label}$, which supervise the point-wise heatmap on point clouds. Eventually, the total loss is formulated as:

\begin{equation}
\label{Equ:loss}
    \mathbf{\mathcal{L}}_{total} = \lambda_{1}\mathbf{\mathcal{L}}_{CE} + \lambda_{2}\mathbf{\mathcal{L}}_{KL} + \lambda_{3}\mathbf{\mathcal{L}}_{HM},
\end{equation}
where $\lambda_1$, $\lambda_2$ and $\lambda_3$ are hyper-parameters to balance the total loss. See more details in supplementary materials.

\begin{table*}[!t]
\centering
\small
  \renewcommand{\arraystretch}{1.}
  \renewcommand{\tabcolsep}{4.pt}
  \caption{{\textbf{Comparison Results on PIAD.} The overall results of all comparative methods, the best results are in \textbf{bold}. \textbf{\texttt{Seen}} and \textbf{\texttt{Unseen}} are two partitions of the dataset. ``\textbf{\textcolor[rgb]{0.2,0.8,0.1}{Green}}" and ``\textbf{\textcolor[rgb]{0.99,0.5,0.0}{Orange}}" indicate two types of the comparative modular baselines. ``\textbf{\texttt{Base.}}'' represents the baseline.  $\textcolor{darkpink}{\diamond}$ denotes the relative improvement of our method over other methods. AUC and aIOU are shown in percentage.}}
\label{table:results}
\vspace{5pt}
\begin{tabular}{c|r|c|ccc|ccc|c}
\toprule
 & \textbf{Metrics} & \textbf{\texttt{Base.}} & \textbf{\textcolor[rgb]{0.2,0.8,0.1}{\texttt{MBDF}} \cite{tan2021mbdf}} & \textbf{\textcolor[rgb]{0.2,0.8,0.1}{\texttt{PMF}} \cite{zhuang2021perception}} & \textbf{\textcolor[rgb]{0.2,0.8,0.1}{\texttt{FRCNN}} \cite{xu2022fusionrcnn}} & \textbf{\textcolor[rgb]{0.99,0.5,0.0}{\texttt{ILN}} \cite{chen2022imlovenet}} & \textbf{\textcolor[rgb]{0.99,0.5,0.0}{\texttt{PFusion}} \cite{xu2018pointfusion}} & \textbf{\textcolor[rgb]{0.99,0.5,0.0}{\texttt{XMF}} \cite{aiello2022cross}}  & \textbf{Ours}  \\
\midrule
\multirow{4}{*}{\rotatebox{90}{\textbf{\texttt{Seen}}}}   & AUC $\uparrow$     & $68.49$    & $74.88\textcolor{darkpink}{\diamond13.3\%}$    & $75.05\textcolor{darkpink}{\diamond13.0\%}$     &  $76.05\textcolor{darkpink}{\diamond11.5\%}$   & $75.84\textcolor{darkpink}{\diamond11.9\%}$      & $77.50\textcolor{darkpink}{\diamond9.5\%}$       & $78.24\textcolor{darkpink}{\diamond8.4\%}$  & \cellcolor{mygray}\textbf{84.85 $\pm$ 0.3} \\
        & aIOU $\uparrow$   & $6.85$     & $9.34\textcolor{darkpink}{\diamond119.5\%}$        & $10.13\textcolor{darkpink}{\diamond102.4\%}$     & $11.97\textcolor{darkpink}{\diamond71.3\%}$   & $11.52\textcolor{darkpink}{\diamond78.0\%}$      & $12.31\textcolor{darkpink}{\diamond66.6\%}$       & $12.94\textcolor{darkpink}{\diamond58.5\%}$ & \cellcolor{mygray}\textbf{20.51 $\pm$ 0.7} \\
        & SIM $\uparrow$    & $0.367$    & $0.415\textcolor{darkpink}{\diamond31.3\%}$        & $0.425\textcolor{darkpink}{\diamond28.2\%}$     & $0.429\textcolor{darkpink}{\diamond27.0\%}$    & $0.427\textcolor{darkpink}{\diamond27.6\%}$       & $0.432\textcolor{darkpink}{\diamond26.1\%}$        & $0.441\textcolor{darkpink}{\diamond23.6\%}$ & \cellcolor{mygray}\textbf{0.545 $\pm$ 0.02} \\
        & MAE $\downarrow$    & $0.152$    & $0.143\textcolor{darkpink}{\diamond31.4\%}$        & $0.141\textcolor{darkpink}{\diamond30.5\%}$      & $0.136\textcolor{darkpink}{\diamond27.9\%}$    & $0.137\textcolor{darkpink}{\diamond28.4\%}$      & $0.135\textcolor{darkpink}{\diamond27.4\%}$       & $0.127\textcolor{darkpink}{\diamond22.8\%}$   & \cellcolor{mygray}\textbf{0.098 $\pm$ 0.01}  \\
\midrule
\multirow{4}{*}{\rotatebox{90}{\textbf{\texttt{Unseen}}}}  & AUC $\uparrow$    & $57.34$    & $58.23\textcolor{darkpink}{\diamond23.3\%}$    & $60.25\textcolor{darkpink}{\diamond19.2\%}$      & $61.92\textcolor{darkpink}{\diamond16.0\%}$    & $59.69\textcolor{darkpink}{\diamond20.3\%}$      & $61.87\textcolor{darkpink}{\diamond16.1\%}$       & $62.58\textcolor{darkpink}{\diamond14.8\%}$ & \cellcolor{mygray}\textbf{71.84 $\pm$ 1.8}     \\
        & aIOU $\uparrow$   & $3.95$    & $4.22\textcolor{darkpink}{\diamond88.4\%}$         & $4.67\textcolor{darkpink}{\diamond70.2\%}$      & $5.12\textcolor{darkpink}{\diamond55.2\%}$     & $4.71\textcolor{darkpink}{\diamond68.8\%}$       & $5.33\textcolor{darkpink}{\diamond49.1\%}$        & $5.68\textcolor{darkpink}{\diamond39.9\%}$  & \cellcolor{mygray}\textbf{7.95 $\pm$ 0.8}     \\
        & SIM $\uparrow$    & $0.318$    & $0.325\textcolor{darkpink}{\diamond8.3\%}$        & $0.330\textcolor{darkpink}{\diamond6.6\%}$     & $0.332\textcolor{darkpink}{\diamond6.0\%}$    & $0.325\textcolor{darkpink}{\diamond8.3\%}$      & $0.330\textcolor{darkpink}{\diamond6.6\%}$       & $0.342\textcolor{darkpink}{\diamond2.9\%}$ & \cellcolor{mygray}\textbf{0.352 $\pm$ 0.03}     \\
        & MAE $\downarrow$    & $0.235$    & $0.213\textcolor{darkpink}{\diamond40.3\%}$        & $0.211\textcolor{darkpink}{\diamond39.8\%}$     & $0.195\textcolor{darkpink}{\diamond34.8\%}$    & $0.207\textcolor{darkpink}{\diamond38.6\%}$      & $0.193\textcolor{darkpink}{\diamond34.2\%}$       & $0.188\textcolor{darkpink}{\diamond32.4\%}$ & \cellcolor{mygray}\textbf{0.127 $\pm$ 0.01}   \\
\bottomrule
\end{tabular}
\end{table*}
\section{Dataset}
\label{sec:dataset}
\par \textbf{Collection Details.} We collect \textbf{P}oint-\textbf{I}mage \textbf{A}ffordance \textbf{D}ataset (\textbf{PIAD}), which contains paired image-point cloud affordance data. The point clouds are mainly collected from 3D-AffordanceNet \cite{deng20213d}, including the point cloud coordinates and affordance annotation. Images are mainly collected from HICO \cite{chao2015hico}, AGD20K \cite{luo2022learning}, and websites with free licenses. The collection criteria is that the image should demonstrate interactions that the object in the point cloud could afford. For example, if the object point cloud is a ``Chair'', it affords ``Sit'', then the image should depict a subject (usually humans) sitting on a chair. The final dataset comprises 7012 point clouds and 5162 images, spanning 23 object classes and 17 affordance categories. Notably, objects in the images and point clouds do not sample from the same physical instance, but they belong to the same object category. Paired examples are shown in Fig. \ref{Fig:dataset} (a).

\par \textbf{Annotation Details.} For point clouds, each point may afford one or multiple interactions, \eg one annotation is a matrix of (2048, 17), $2048$ is the point number, $17$ represents the number of affordance types. Each element in the matrix indicates the probability of a point affording a specific affordance. Meanwhile, images are annotated with bounding boxes of the interactive subject and object, as well as an affordance category label. In our task, we only use the heatmap in the point cloud annotation that corresponds to the affordance of the image for training, resulting in a matrix of (2048, 1). By doing so, the affordance category is detached from the point cloud during inference, and the anticipation of the 3D object affordance category only relies on 2D interactions. As a result, different affordances can be anticipated on a point cloud through distinct interactions.

\par \textbf{Statistic Analysis. } Since images and point clouds are sampled from different instances, they do not need a fixed one-to-one pairing, one image could be paired with multiple point clouds and the count of them is not strictly consistent. Fig. \ref{Fig:dataset} (b) and (c) show the count and distribution of affordances in images and point clouds. Fig. \ref{Fig:dataset} (d) illustrates the ratio of images and point clouds in each affordance category. PIAD has two partitions: \textbf{\texttt{Seen}} and \textbf{\texttt{Unseen}}. In \textbf{\texttt{Seen}}, both objects and affordances in the training and testing sets are consistent, while in \textbf{\texttt{Unseen}}, some objects or affordances in the testing set do not exist in the training set.
\section{Experiments}
\subsection{Benchmark Setting}
\label{5.1}
\noindent\textbf{Evaluation Metrics.\ } To provide a comprehensive and effective evaluation, we compare serval advanced works in the affordance area \cite{deng20213d, nagarajan2019grounded, zhai2022one} and finally chose four evaluation metrics: \textbf{AUC} \cite{lobo2008auc}, \textbf{aIOU} \cite{rahman2016optimizing}, \textbf{SIM}ilarity \cite{swain1991color} and \textbf{M}ean \textbf{A}bsolute \textbf{E}rror \cite{willmott2005advantages} to benchmark the PIAD.

\noindent\textbf{Modular Baselines.\ } Since there are no prior works using paired image-point cloud data to ground 3D object affordance. For a thorough comparison of our method, we select several advanced image-point cloud cross-modal works as modular baselines. Methods of comparison are broadly divided into two types, one is that utilizes camera intrinsics to facilitate cross-modal feature alignment or fusion, \ie \textbf{\textcolor[rgb]{0.2,0.8,0.1}{\texttt{MBDF-Net(MBDF)}}} \cite{tan2021mbdf}, \textbf{\textcolor[rgb]{0.2,0.8,0.1}{\texttt{PMF}}} \cite{zhuang2021perception} and \textbf{\textcolor[rgb]{0.2,0.8,0.1}{\texttt{FusionRCNN(FRCNN)}}} \cite{xu2022fusionrcnn}, for this type of method, we remove the step of using intrinsic parameters to align raw data or features to explore their effectiveness on PIAD. Another type performs feature alignment directly in the feature space, without relying on camera intrinsic, \ie \textbf{\textcolor[rgb]{0.99,0.5,0.0}{\texttt{ImLoveNet(ILN)}}} \cite{chen2022imlovenet}, \textbf{\textcolor[rgb]{0.99,0.5,0.0}{\texttt{PointFusion(PFusion)}}} \cite{xu2018pointfusion} and \textbf{\textcolor[rgb]{0.99,0.5,0.0}{\texttt{XMFnet(XMF)}}} \cite{aiello2022cross}. To ensure a fair comparison, all methods use the same feature extractors, with the only variation coming from cross-modal alignment or fusion block. For the \textbf{\texttt{Baseline}}, we directly concatenate the features that are output from extractors, regarding the concatenation as the cross-fusion block. More details about baselines are provided in the supplementary materials.

\begin{figure}[t]
    \centering
    \footnotesize
    \begin{overpic}[width=0.96\linewidth]{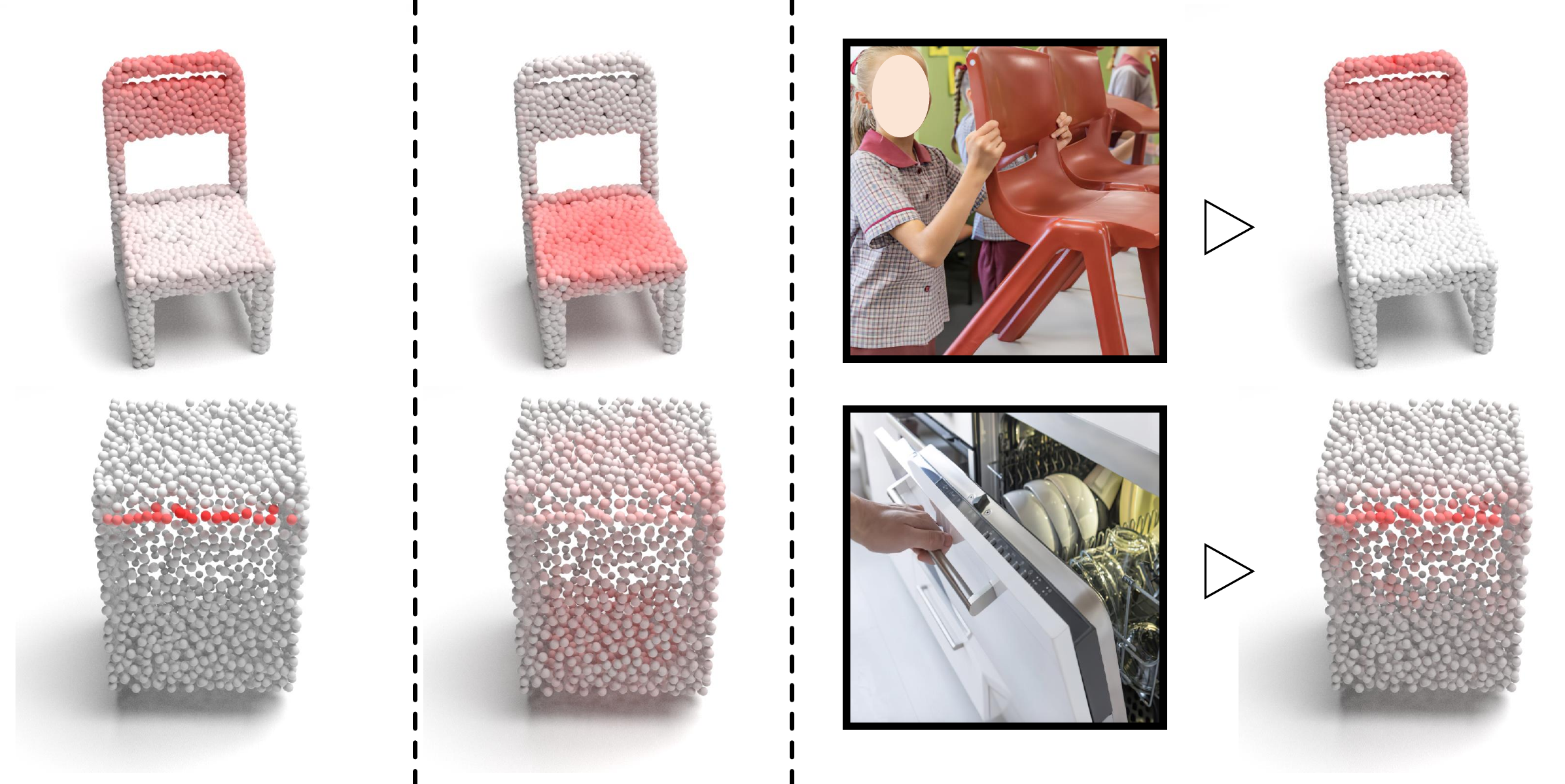}
    \put(11,0){\textbf{(a)}}
    \put(10,49){\textbf{GT}}
    \put(76,0){\textbf{(c)}}
    \put(75,49){\textbf{Ours}}
    \put(36.5,0){\textbf{(b)}}
    \put(36,49){\textbf{\cite{deng20213d}}}
    \put(-1,33){\rotatebox{90}{\textbf{Move}}}
    \put(-1,9){\rotatebox{90}{\textbf{Open}}}
    \end{overpic}
    \caption{\textbf{Paradigm Comparison.} \textbf{(a)} Ground truth. \textbf{(b)} The results of 3D-AffordanceNet \cite{deng20213d}. \textbf{(c)} Our results. The top row shows the regional confusion case in \textbf{\texttt{Seen}}, and the bottom row displays the generalization in \textbf{\texttt{Unseen}}.}
    \label{fig:only_point}
\end{figure}

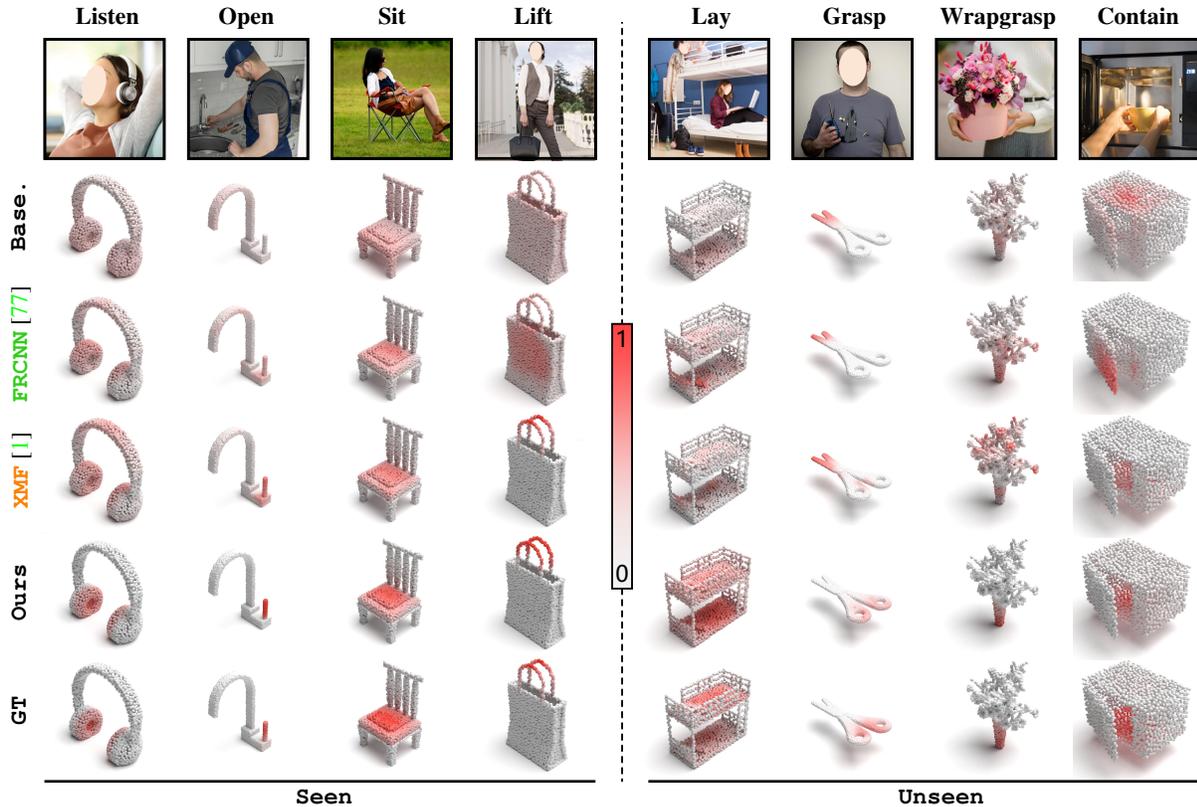
\begin{figure*}[t]
	\centering
	\small
        \begin{overpic}[width=0.9\linewidth]{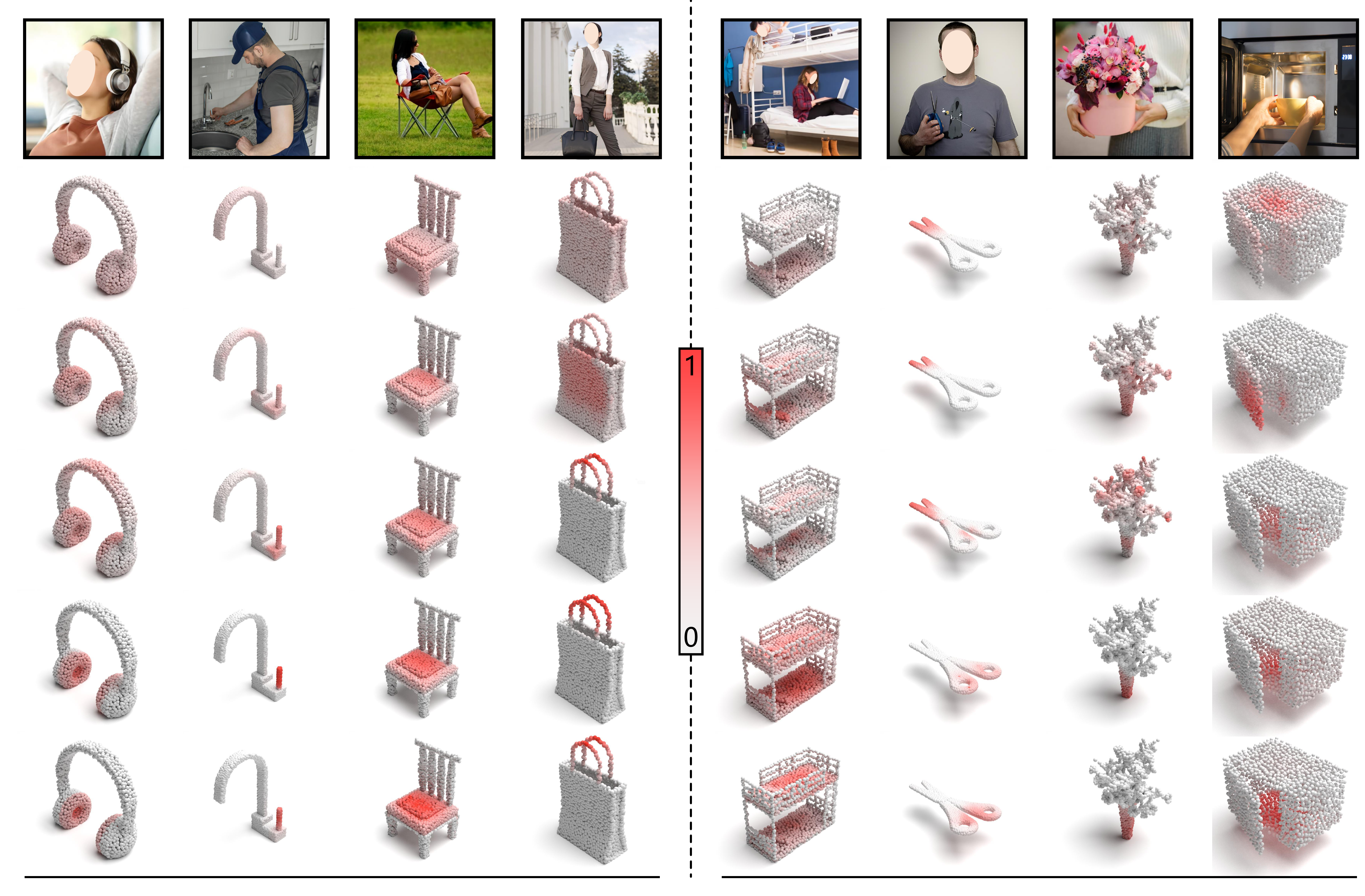}
        \put(23,-0.5){{\textbf{\texttt{Seen}}}}
        \put(74,-0.5){{\textbf{\texttt{Unseen}}}}
        
        \put(4.4,65.5){{\textbf{Listen}}}
        \put(16.5,65.5){{\textbf{Open}}}
        \put(30,65.5){{\textbf{Sit}}}
        \put(41.5,65.5){{\textbf{Lift}}}
        \put(56.5,65.5){{\textbf{Lay}}}
        \put(67.6,65.5){{\textbf{Grasp}}}
        \put(77.5,65.5){{\textbf{Wrapgrasp}}}
        \put(90.8,65.5){{\textbf{Contain}}}
        
        \put(-1,6.2){\rotatebox{90}{\textbf{\texttt{GT}}}}
        \put(-1,15){\rotatebox{90}{\textbf{\texttt{Ours}}}}
        \put(-1,24.5){\rotatebox{90}{\textcolor[rgb]{0.99,0.5,0.0}{\textbf{\texttt{XMF}}} \cite{aiello2022cross}}}
        \put(-1,33.5){\rotatebox{90}{\textcolor[rgb]{0.2,0.8,0.1}{\textbf{\texttt{FRCNN}}} \cite{xu2022fusionrcnn}}}
        \put(-1,46.3){\rotatebox{90}{\textbf{\texttt{Base.}}}}
	\end{overpic}
	\caption{\textbf{Visualization Results.} The first row is the interactive image, which demonstrates the interaction that the object can afford. The last row is the ground truth of 3D object affordance in the point cloud. The four columns on the left are the visual comparison results for distinct 3D object affordances in the \textbf{\texttt{Seen}} partition. The four columns on the right are the results in the \textbf{\texttt{Unseen}} partition.}
 \label{Fig:mainresults}
\vspace{1pt}
\end{figure*}

\subsection{Comparison Results}
The comparison results of evaluation metrics are shown in Tab. \ref{table:results}. As can be seen, our method outperforms the compared baselines, across all metrics in both partitions. Besides, to display the limitation of methods that lock geometrics with specific semantic affordance categories, we conduct an experiment to compare one of these methods \cite{deng20213d} with ours. As shown in Fig. \ref{fig:only_point}, the top row indicates that the result of this method exhibits regional confusion, \eg the region where the chair could be moved or sat is geometrically rectangular, and it directly anticipates results on these similar geometries, inconsistent with affordance property. Plus, the bottom row shows that directly establishing a link between object structures and affordances may fail to anticipate correct 3D affordance in \textbf{\texttt{Unseen}}. In contrast, our method anticipates precise results by mining affordance clues provided by 2D interactions. Additionally, visual comparative results of our method and other baselines are shown in Fig. \ref{Fig:mainresults}. As can be seen, the comparative baselines could anticipate some 3D object affordance under our setting, but in comparison, our method obviously achieves better results, which validates the rationality of our setting, and also demonstrates the superiority of our method.

\begin{table}
\centering
\small
\renewcommand{\arraystretch}{1.}
\renewcommand{\tabcolsep}{7.5pt}
\caption{\textbf{Ablation Study.} We investigate the improvement of JRA and ARM on the model performance based on the baseline.
}
\label{table:ablation}
\vspace{5pt}
\begin{tabular}{c|cc|cccc}
\toprule
\multicolumn{1}{l|}{} & \textbf{JRA} & \textbf{ARM} &\textbf{AUC} & \textbf{aIOU} & \textbf{SIM} & \textbf{MAE} \\ 
\midrule
\multirow{4}{*}{\rotatebox{90}{\textbf{\texttt{Seen}}}} &  &  &  69.92 & 8.85 & 0.427 & 0.132 \\
& \checkmark &  &  80.29 & 14.31 & 0.495 & 0.121 \\
 & & \checkmark & 78.67 & 13.95 & 0.475 & 0.126 \\
 & \checkmark & \checkmark & \cellcolor{mygray}\textbf{85.16} & \cellcolor{mygray}\textbf{21.20} & \cellcolor{mygray}\textbf{0.564} & \cellcolor{mygray}\textbf{0.088} \\ 
 \midrule
\multirow{4}{*}{\rotatebox{90}{\textbf{\texttt{Unseen}}}} & &  & 59.14 & 4.05 & 0.338 & 0.202 \\
& \checkmark &  & 66.25 & 6.27 & 0.363 & 0.159 \\
 & & \checkmark & 65.79 & 5.99 & 0.358 & 0.162 \\
 & \checkmark & \checkmark & \cellcolor{mygray}\textbf{73.69} & \cellcolor{mygray}\textbf{8.70} & \cellcolor{mygray}\textbf{0.383} & \cellcolor{mygray}\textbf{0.117} \\
 \bottomrule
\end{tabular}
\end{table}

\begin{figure}
    \centering
    \footnotesize
    \begin{overpic}[width=0.9\linewidth]{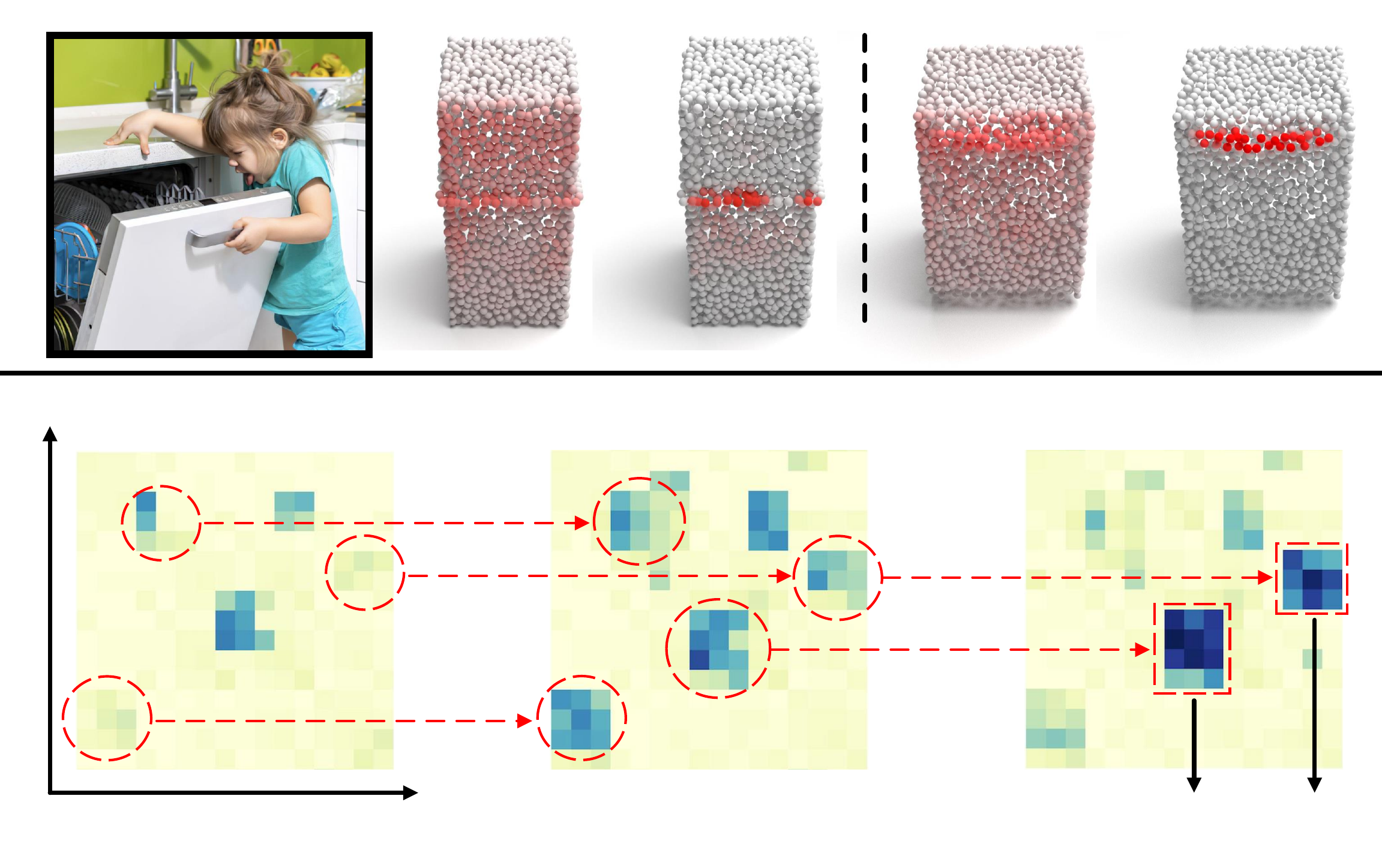}
    \put(5,1){\textbf{Image Regions}}
    \put(-0.5,9){\textbf{\rotatebox{90}{Point Regions}}}
    \put(81,3){\textbf{Affordance}}
    \put(84,-0.5){\textbf{Regions}}
    \put(-0.5,44){\textbf{\rotatebox{90}{Open}}}
    \put(28.5,61){\textbf{$\bm{w/o}$ JRA}}
    \put(48.5,61){\textbf{$\bm{w}$ JRA}}
    \put(65,61){\textbf{$\bm{w/o}$ JRA}}
    \put(85,61){\textbf{$\bm{w}$ JRA}}
    \put(43,32){\textbf{step=8000}}
    \put(10,32){\textbf{step=1000}}
    \put(77,32){\textbf{step=15000}}
    \end{overpic}
    \caption{\textbf{Ablation of JRA.} The top row is the result of the anticipated affordance with/without JRA. The bottom row is a part of the cross-similarity matrix of a sample during the training process.}
    \label{fig:kl}
\end{figure}

\begin{figure}
    \centering
    \footnotesize
    \begin{overpic}[width=0.87\linewidth]{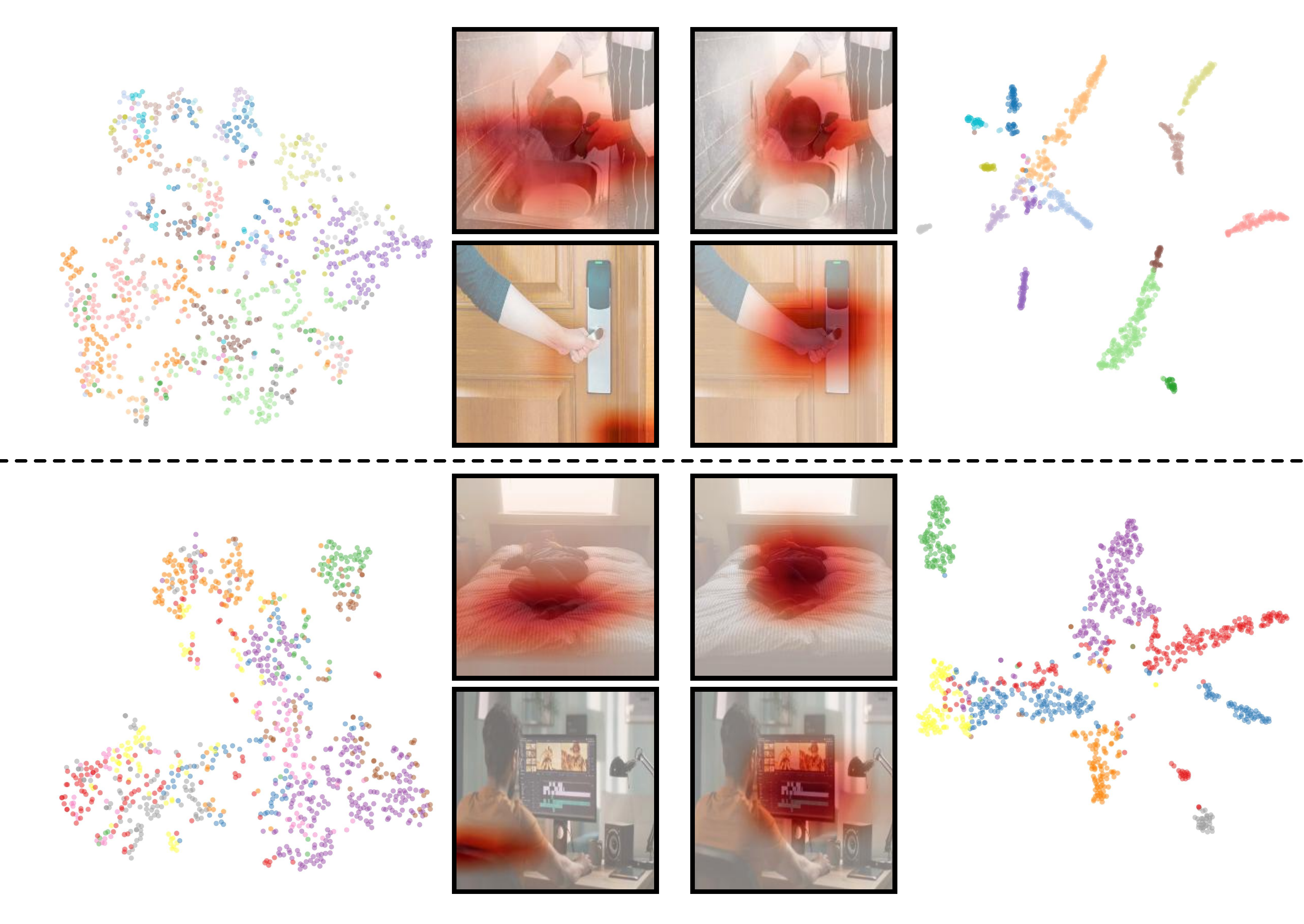}
    \put(20,70){\textbf{$\bm{w/o}$ ARM}}
    \put(65,70){\textbf{$\bm{w}$ ARM}}
    \put(-0.5,50){\rotatebox{90}{\textbf{\texttt{Seen}}}}
    \put(-0.5,10){\rotatebox{90}{\textbf{\texttt{Unseen}}}}
    \end{overpic}
    \caption{\textbf{Ablation of ARM.} The activation map and t-SNE \cite{van2008visualizing} results with ($\bm{w}$), without ($\bm{w/o}$) ARM in both partitions.}
    \label{fig:heatmap}
\end{figure}

\begin{figure}
    \centering
    \small
    \begin{overpic}[width=0.93\linewidth]{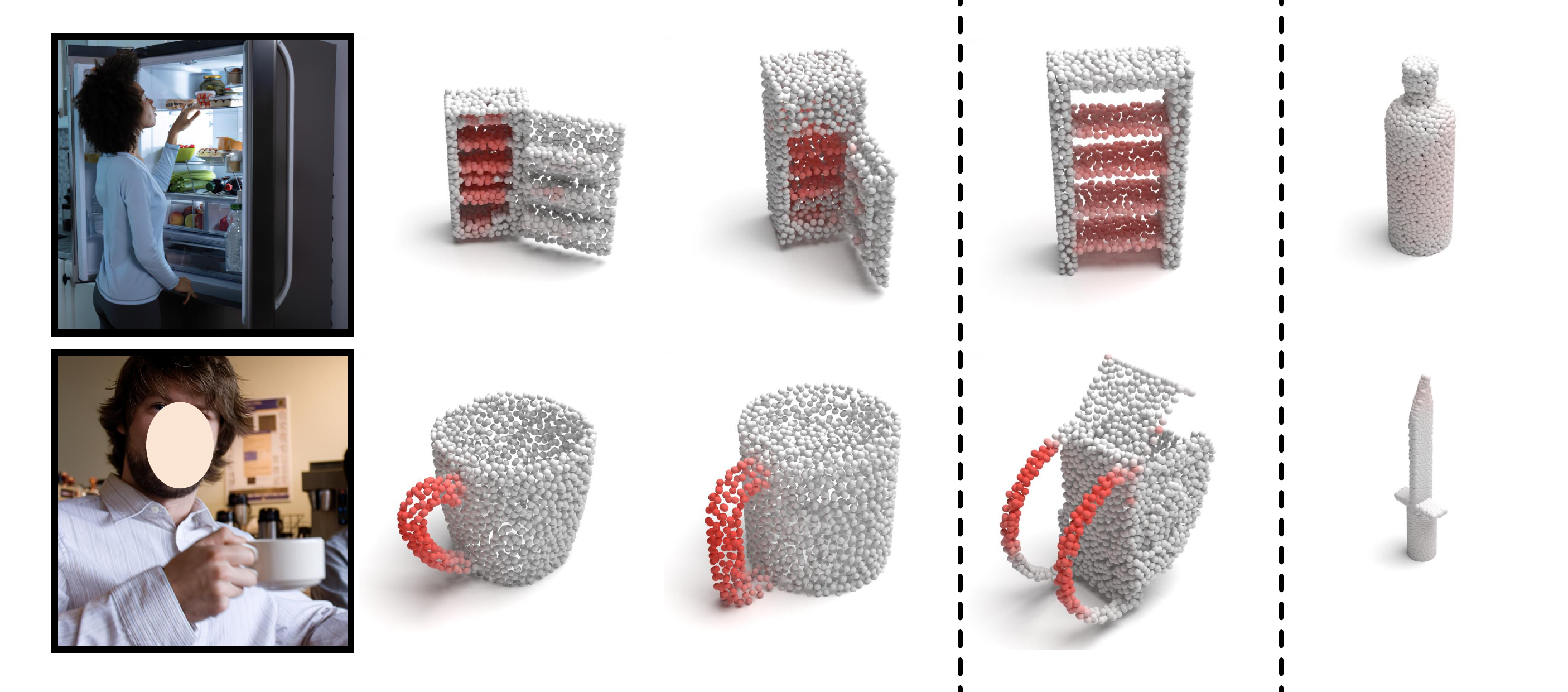}
    \put(39,-0.8){\textbf{(a)}}
    \put(68.5,-0.8){\textbf{(b)}}
    \put(88.5,-0.8){\textbf{(c)}}
    \put(-1,6){\textbf{\rotatebox{90}{Grasp}}}
    \put(-1,24){\textbf{\rotatebox{90}{Contain}}}
    \end{overpic}
    \caption{\textbf{Same Image} $w.r.t.$ \textbf{Multiple Point Clouds.} \textbf{(a)} Same object category. \textbf{(b)} Different object categories, similar geometrics. \textbf{(c)} Different object categories and geometrics.}
    \label{fig:one2multi1}
\end{figure}

\begin{figure}
    \centering
    \small
    \begin{overpic}[width=0.93\linewidth]{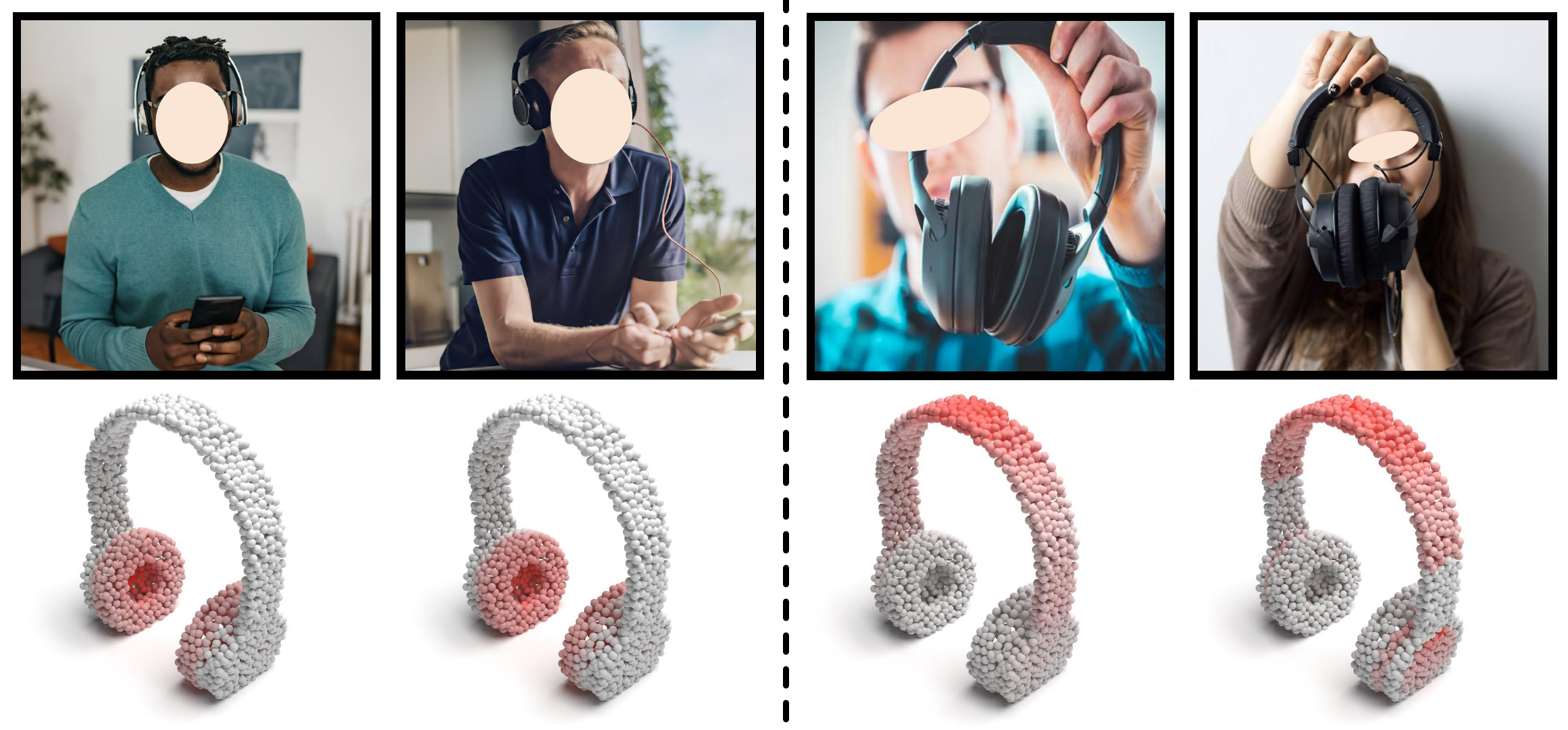}
    \put(20,-1){\textbf{Listen}}
    \put(70,-1){\textbf{Grasp}}
    \end{overpic}
    \caption{\textbf{Same Point Cloud} $w.r.t.$ \textbf{Multiple Images.} Grounding affordance on the same point cloud with images that contain similar or disparate interactions.}
    \label{fig:one2multi2}
\end{figure}

\subsection{Ablation Study}
\label{5.3}
\noindent\textbf{Effectiveness of JRA. } Tab. \ref{table:ablation} reports the impact on
evaluation metrics of JRA. Fig. \ref{fig:kl} provides visual comparison results. It shows that without the JRA, the result is anticipated over the entire region that contains the interactive components, which means the 2D-3D affordance regions of objects do not match well. Besides, we visualize a part of the cross-similarity matrix between $\hat{\mathbf{F}}_{p}$ and $\hat{\mathbf{F}}_{i}$ in Fig. \ref{fig:kl}. In the initial stage, only a few analogous shape regions keep explicit correspondence, as training proceeds, the JRA maps the correlation among regions with lower similarity. Meanwhile, the model extracts explicit affordance progressively, and the affordance introduced into the optimization process reveals the corresponding affordance regions.

\noindent\textbf{Effectiveness of ARM. } The influence of ARM on evaluation metrics is also shown in Tab. \ref{table:ablation}. To visually evaluate the effectiveness of ARM, we employ GradCAM \cite{selvaraju2017grad} to generate the activation map, and utilize t-SNE \cite{van2008visualizing} to demonstrate the clustering of affordances in both partitions. The results are shown in Fig. \ref{fig:heatmap}. It shows that ARM makes the model focus on the interactive regions to excavate the interaction contexts, and enables the model to differentiate various affordances from interactions in both partitions.

\subsection{Performance Analysis}
\label{6.1}
\noindent\textbf{Different Instances.} We conduct experiments to verify whether the model could ground 3D affordance on instances from different sources: (\romannumeral1) using an image and different point clouds to infer respectively (Fig. \ref{fig:one2multi1} (a)). (\romannumeral2) using multiple images and a single point cloud (Fig. \ref{fig:one2multi2}). The results indicate that the model can anticipate 3D affordances on different instances with the same 2D interaction, and can also anticipate distinct 3D affordances on the same point cloud through different 2D interactions. Showing that the model maintains the mapping between object structures and affordances by learning from different 2D interactions.

\noindent\textbf{Different Object Categories.} What will happen if the object category is different in the image and point cloud? To explore this issue, we perform experiments with mismatched object categories, shown in Fig. \ref{fig:one2multi1} (b) and (c). When objects are of different categories but with similar geometric primitives, \eg the handle of ``Mug'' and the shoulder strap of ``Bag'', our model still properly infer the 3D affordance. However, when geometric structures are also dissimilar (Fig. \ref{fig:one2multi1} (c)), the model does not make random predictions. This indicates that the model maps cross-category invariance between affordance and geometries, which can be generalized to new instances of geometry.

\begin{figure}
    \centering
    \footnotesize
    \begin{overpic}[width=0.9\linewidth]{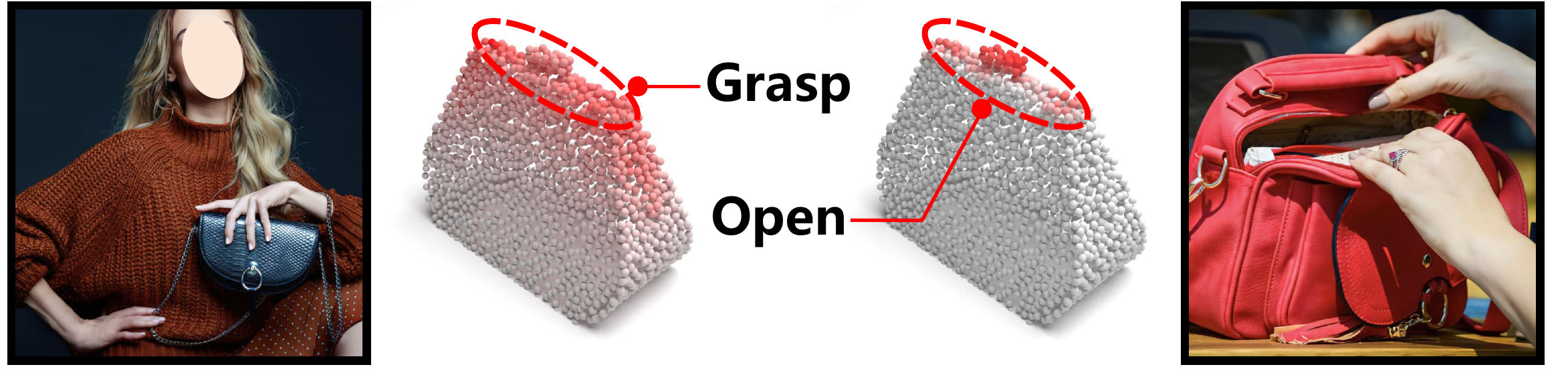}
    \put(-3.5,7){\textbf{\rotatebox{90}{Grasp}}}
    \put(99.5,7){\textbf{\rotatebox{90}{Open}}}
    \end{overpic}
    \caption{\textbf{Multiple Affordances.} Some objects like ``Bag" contains the region that corresponds to multiple affordances.}
    \label{fig:multi_property}
\end{figure}

\noindent\textbf{Multiplicity. } One defining property of affordance is multiplicity: some points may correspond to multiple affordances. Fig. \ref{fig:multi_property} shows that the model makes objects' multiple affordances compatible, and verifies the model does not anticipate affordance by linking a region with specific affordances. Instead, it predicts by considering the presence of affordance-related interactions in the same region.

\begin{figure}
	\centering
        \footnotesize
		\begin{overpic}[width=0.95\linewidth]{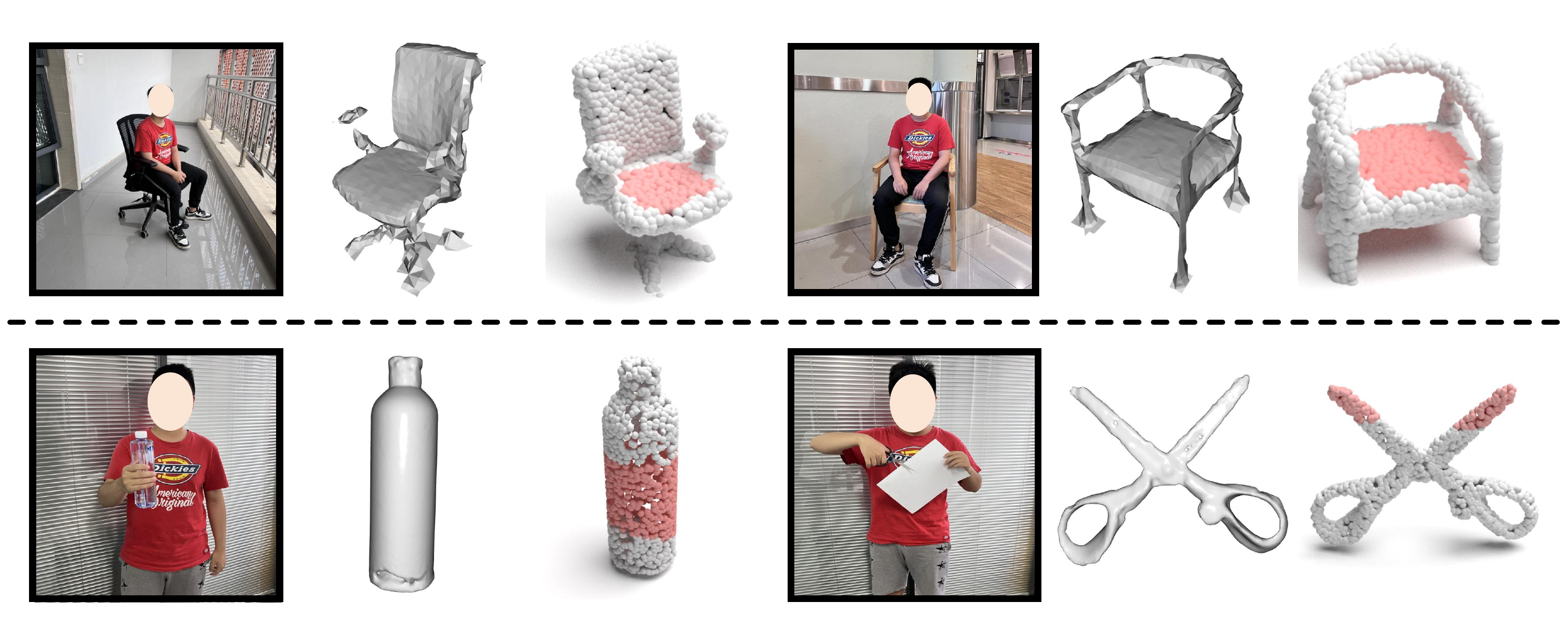}
		\put(6,39){\textbf{Image}}
		\put(23,39){\textbf{Scan}}
		\put(36,39){\textbf{Predict}}
  		\put(54,39){\textbf{Image}}
		\put(70.5,39){\textbf{Scan}}
		\put(84,39){\textbf{Predict}}
		\put(-2,27.5){\rotatebox{90}{\textbf{Sit}}}
  		\put(-2,0.6){\rotatebox{90}{\textbf{Wrapgrasp}}}
  		\put(46.8,7.3){\rotatebox{90}{\textbf{Cut}}}

	\end{overpic}
	\caption{\textbf{Real-Wrold.} The first row is scanned by iPhone, the second row comes from \cite{Liu_2022_CVPR}, point clouds are sampled from them.}
 \label{fig:real}
\end{figure}

\noindent\textbf{Real World. } To validate the model in real-world scenarios, we use an iPhone to scan objects and real-scan dataset \cite{Liu_2022_CVPR} to test our model, shown in Fig. \ref{fig:real}. It shows that our model exhibits a certain degree of generalization to the real world.

\begin{figure}[t]
    \centering
    \footnotesize
    \begin{overpic}[width=0.96\linewidth]{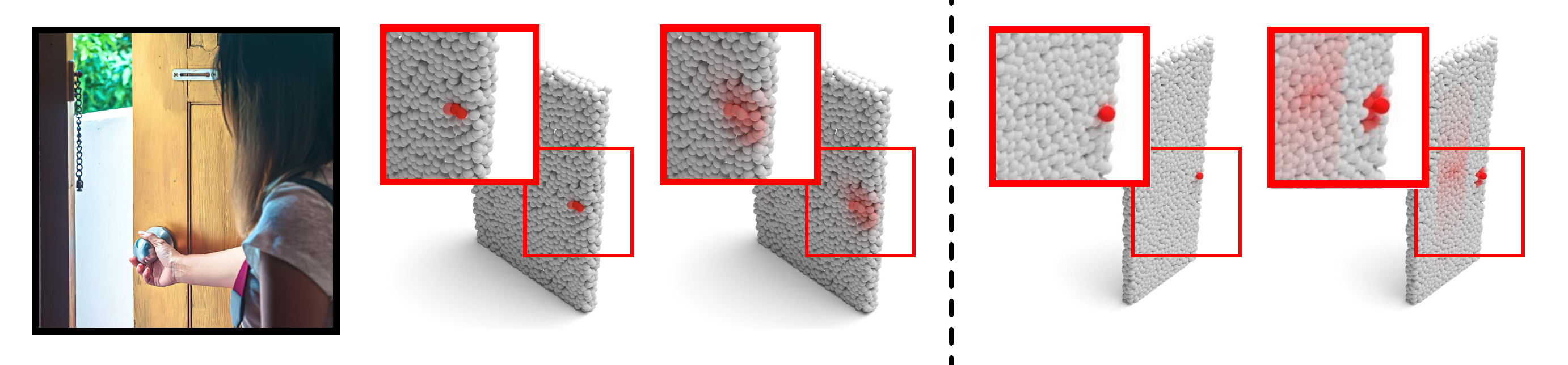}
    \put(8,23.5){\textbf{Open}}
    \put(31,23.5){\textbf{GT}}
    \put(46,23.5){\textbf{Ours}}
    \put(70,23.5){\textbf{GT}}
    \put(86,23.5){\textbf{Ours}}
    \end{overpic}
    \caption{\textbf{Failure Cases.}  Over prediction in the small regions.}
    \label{fig:limitation}
\end{figure}
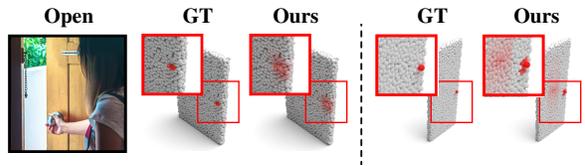

\noindent\textbf{Limitations.} Our model exhibits over predictions for some small affordance regions, shown in Fig. \ref{fig:limitation}. This may be due to the limited understanding of fine-grained geometric parts, and aligning the object features in a single scale may lead to overly large receptive fields. Our feature extension will refer \cite{Geng_2023_CVPR, lin2017feature} to tackle this problem.
\section{Conclusion}
We present a novel setting for affordance grounding, which utilizes the 2D interactive semantics to guide the grounding of 3D object affordance, it has the potential to serve embodied systems when collaborating with multi-modal grounding systems \cite{liu2019adaptive, tan2020learning, yang2019making}. Besides, We collect the PIAD dataset as the first test bed for the proposed setting, it contains paired image-point cloud affordance data. Plus, we propose a novel framework to correlate affordance regions of objects that are from different sources and model interactive contexts to ground 3D object affordance. Comprehensive experiments on PIAD display the reliability of the setting, and we believe it could offer fresh insights and facilitate research in the affordance area. \\
\textbf{Acknowledgments} This work is supported by National Key R\&D Program of China under Grant 2020AAA0105700, National Natural Science Foundation of China (NSFC) under Grants 62225207, U19B2038 and 62121002.

{\small
\bibliographystyle{ieee_fullname}
\bibliography{egbib}
}
\clearpage

\section{Supplementary Material}
\section*{A. Implementation Details}
\addcontentsline{toc}{section}{\textcolor[rgb]{0,0,0}{A. Implementation Details}}

\subsection*{A.1. Method Details}
\addcontentsline{toc}{subsection}{\textcolor[rgb]{0,0,0}{A.1. Method Details}}

\begin{figure}[h]
	\centering
        \small
		\begin{overpic}[width=1\linewidth]{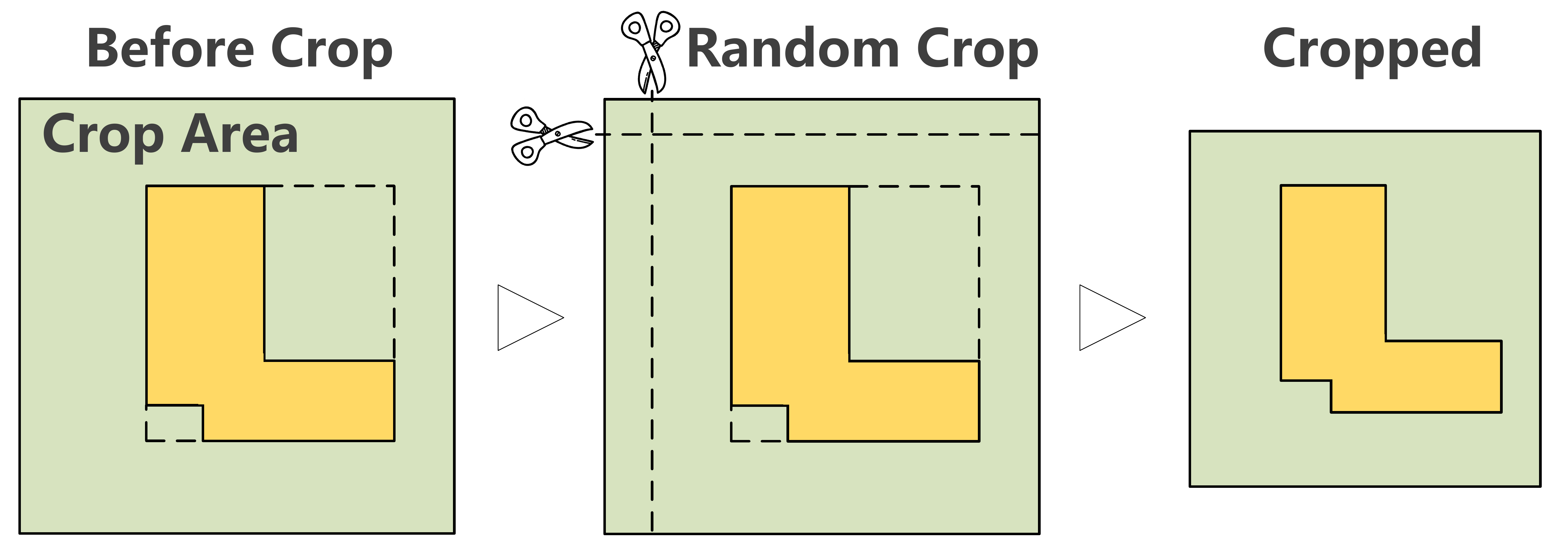}

	\end{overpic}
	\caption{\textbf{Random Crop.} Cropping in image augmentation, we only do it outside the object and subject bounding boxes.}
	\label{fig:Crop}
\end{figure}

For the image branch, we chose ResNet18 as the extractor, and the input images are randomly cropped and resized to 224$\times$224. To prevent the interactive subject and object in the image from being cropped out, we set the crop area outside the bounding box of the interactive subject and object, shown in Fig. \ref{fig:Crop}. The image extractor output the image feature $\mathbf{F_I} \in \mathbb{R}^{512 \times 7 \times 7}$. Then, utilizing the bounding boxes $B_{obj}, B_{sub}$ to calculate the scene mask $M_{sce}$ (outside these two boxes), taking them to locate the object, subject, and scene feature in $\mathbf{F_I}$, next, apply Roi-Align to obtain the same size object, subject, scene feature $\mathbf{F}_{i}, \mathbf{F}_{s}, \mathbf{F}_{e} \in \mathbb{R}^{512 \times 4 \times 4}$, reshape them to $\mathbb{R}^{512 \times 16}$. For the point cloud branch, the number of points for each input point cloud is fixed to $2048$, we take $3$ set abstraction (SA) layers with multi-scale grouping to extract the point-wise feature. Each SA layer uses Farthest Point Strategy (FPS) to sample points, and the number of sampling points for each layer is set to 512, 128, and 64 respectively. Finally, this branch outputs the point-wise feature $\mathbf{F}_{p} \in \mathbb{R}^{512 \times 64}$. We show the dimension of tensors in the whole pipeline in Tab. \ref{tab:tensor}. In the implementation, the joint denotes combining the image and point cloud feature sequence at the last dimension, like joint $\bar{\mathbf{P}}$ and $\bar{\mathbf{I}}$ as the object representation $\mathbf{F}_{j}$.

\begin{table}[]
  \centering
  \renewcommand{\arraystretch}{1.}
  \renewcommand{\tabcolsep}{8pt}
    \footnotesize
   \caption{\textbf{Tensors.} The dimension and meaning of the tensors in the pipeline.}
   \label{tab:tensor}
   \vspace{5pt}
\begin{tabular}{c|c|c}
\toprule
\textbf{Tensor}    & \textbf{Dimension} & \textbf{Meaning} \\ \midrule
$\mathbf{F_I}$       & $512 \times 7 \times 7$           & image extractor output            \\
$\mathbf{F}_p$    & $512 \times 64$           & point cloud extractor output    \\
$\mathbf{F}_{i,s,e}$        & $512 \times 16$                 & features output by roi-align           \\
$\mathbf{P}$  & $512 \times 64$   & project $\mathbf{F}_p$ to a feature space            \\
$\mathbf{I}$    & $512 \times 16$           & project $\mathbf{F}_i$ to a feature space    \\
$\varphi$  & $64 \times 16$   & dense similarity between $\mathbf{P}$ and $\mathbf{I}$     \\
$\bar{\mathbf{P}}$     &  $512 \times 64$      & point feature with structural relevance           \\
$\bar{\mathbf{I}}$       & $512 \times 16$          & image feature with structural relevance            \\
$\mathbf{F}_j$   & $512 \times 80$               & joint object representation            \\
 $\mathbf{Q}$           &  $512 \times 80$                & the query projected by $\mathbf{F}_j$            \\
$\mathbf{K}_1, \mathbf{K}_2$        & $512 \times 16$             & keys projected by $\mathbf{F}_s, \mathbf{F}_e$            \\
$\mathbf{V}_1, \mathbf{V}_2$        & $512 \times 16$             & values projected by $\mathbf{F}_s, \mathbf{F}_e$            \\
 $\mathbf{\Theta}_1, \mathbf{\Theta}_2$ &     $512 \times 80$  &  interaction contexts \\
  $\mathbf{F}_{\alpha}$ &     $512 \times 80$  &  joint affordance representation \\
$\hat{\mathbf{F}}_{p}, \mathbf{F}_{p\alpha}$        & $512 \times 64$   & split from $\mathbf{F}_j$, $\mathbf{F}_{\alpha}$          \\
$\hat{\mathbf{F}}_{i}, \mathbf{F}_{i\alpha}$        & $512 \times 16$   & split from $\mathbf{F}_j$, $\mathbf{F}_{\alpha}$          \\
$\hat{\phi}$      & $2048 \times 1$                & 3D object affordance            \\ \bottomrule
\end{tabular}
\end{table}

\begin{figure}[t]
	\centering
        \small
		\begin{overpic}[width=0.95\linewidth]{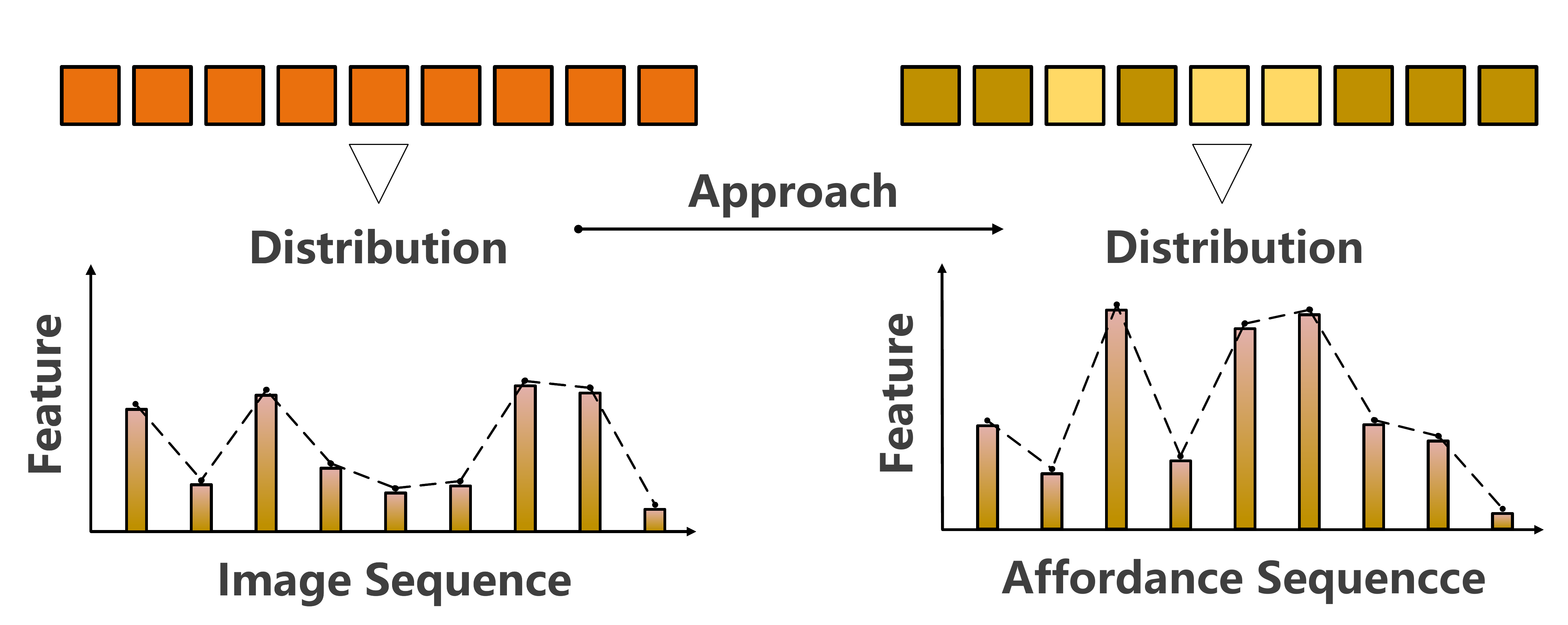}
            \put(23,38){$\hat{\mathbf{F}}_{i}$}
            \put(76,38){$\mathbf{F}_{i\alpha}$}
	\end{overpic}
	\caption{\textbf{The Distribution.} The right denotes feature distribution of each region in $\mathbf{F}_{i\alpha}$, \textbf{\textcolor[RGB]{250,217,115}{regions}} with higher correspondence to affordance keep more significant feature. And the goal is to let this property exists in $\hat{\mathbf{F}}_{i}$, so we narrow the distribution discrepancy between $\hat{\mathbf{F}}_i$ and $\mathbf{F}_{i\alpha}$ to tune the feature distribution of $\hat{\mathbf{F}}_{i}$.}
	\label{fig:KL}
\end{figure}

Furthermore, here we make a more detailed explanation for the KLD loss $\mathcal{L}_{KL}$. $\mathbf{F}_{j}$ denotes the joint object representation, and $\mathbf{F}_{\alpha}$ denotes the joint affordance representation. Since the order of the sequence does not change in the calculation process, we split them back into image and point cloud sequences. $\mathbf{F}_{i\alpha}$ contains the affordance feature distribution of each region in $\hat{\mathbf{F}}_i$, the regions with high correspondence to affordance possess more significant features. This relative difference is reflected in the region feature distribution, the insight is to make the distribution of $\hat{\mathbf{F}}_i$ also keep the distribution characteristics of the $\mathbf{F}_{\alpha}$, so as to implicitly enhance the affordance region features in the object representation, shown in Fig. \ref{fig:KL}. And with the establishment of the correspondence between the image and point cloud regions in the alignment process, this property tends to be shown in $\hat{\mathbf{F}}_p$. This makes the region alignment and affordance extraction exhibit a mutual mechanism, the affordance representation could better correspond the object affordance regions, and the object region features with stronger correspondences could assist to extract more explicit affordance features in the optimization process. The computation of $\mathcal{L}_{KL}$ is expressed as:
\begin{equation}
\small
   \mathcal{L}_{KL} = KLD(\hat{\mathbf{F}}_{i},\mathbf{F}_{i\alpha})=\sum_{n}\mathbf{F}_{i\alpha_{n}}log(\epsilon + \frac{\mathbf{F}_{i\alpha_{n}}}{\epsilon+\hat{\mathbf{F}}_{i_{n}}}),
\end{equation}
where $\epsilon$ is a regularization constant, $n$ denotes the regions. Since $\mathbf{F}_{i\alpha}$ is split from $\mathbf{F}_{\alpha}$ and $\hat{\mathbf{F}}_{i}$ is split from $\mathbf{F}_{j}$, $\mathcal{L}_{KL}$ could optimize the layers for alignment in JRA, also the layers for affordance extraction in ARM, expressed as:
\begin{equation}
    \theta_{1}^{t+1} \leftarrow  \theta_{1}^{t} - \eta \nabla \frac{\partial \mathcal{L}_{KL}(\theta_{1}^{t+1})}{\partial \theta_{1}^{t+1}},
\end{equation}
where $t$ is the number of steps, $\eta$ is the learning rate, $\theta_{1}$ denotes the parameters of layers that tend to be optimized.

\par In JRA, for the learnable layers, we give the following explanations. $f_{\delta}$ contains two $1\times1$ convolution layers, which is used to project $\mathbf{F}_{i}$ and $\mathbf{F}_{p}$ into a feature space. $f_i, f_p$ are utilized to map the region relevance of object feature from images and point clouds. After the feature extraction, the $\mathbf{F}_{p}$ and $\mathbf{F}_{i}$ represent sub-regions of the raw objects, our aim is to map the spatial correlation among these sub-regions. There are several ways to satisfy this mapping: 1) Transformer-based. In this method, each sub-region feature can be regarded as a patch, then take the self-attention technique to model the correlation among the patches. 2) Expectation-Maximization Attention \cite{li2019expectation}. This method initializes a group of bases $\mu$, and then performs alternating ``E-step'' and ``M-step'' to mine the correspondence between object features from the image and point cloud. 3) Multilayer perception. This is the most direct way, which regards all regions as a whole one and directly maps the intra-relation among the region features. We conduct a comparative experiment to see their performance and finally chose the self-attention technique (see Tab. \ref{table:projection_method}).

\par In ARM, for the extraction of affordance representation $\mathbf{F}_{\alpha}$, we take the feature $\mathbf{F}_{j}$ as the object representation to model the interaction contexts and reveal affordance. This process could be regarded as fusing the image and point cloud feature as the multi-modal representation and completing certain downstream tasks like affordance extraction. And the feature alignment is performed implicitly during optimizing the extraction, which is a learning-based way \cite{baltruvsaitis2018multimodal}. Meanwhile, theoretically, the multi-modal feature contains more information, for example, structures, colors, and textures, which is also beneficial to the downstream task like the extraction of affordance (see clarification in \cite{huang2021makes}).

\subsection*{A.2. Evaluation Metrics}
\addcontentsline{toc}{subsection}{\textcolor[rgb]{0,0,0}{A.2. Evaluation Metrics}}
We use four evaluation metrics to benchmark the PIAD. \textbf{AUC} \cite{lobo2008auc} is used to evaluate the predicted saliency map on the point cloud. The \textbf{a}verage \textbf{I}ntersection \textbf{O}ver \textbf{U}nion (\textbf{aIoU}) \cite{rahman2016optimizing} is aimed at evaluating the overlap between the affordance region predicted in the point cloud and the labeled region. \textbf{SIM}ilarity (\textbf{SIM}) \cite{swain1991color} is used to measure the similarity between the prediction map and the ground truth. \textbf{M}ean \textbf{A}bsolute \textbf{E}rror (\textbf{MAE}) \cite{willmott2005advantages} is the absolute difference between the prediction map and ground truth for point-wise measurement.

\begin{itemize}

\item [$\bm{-}$] \textbf{AUC }\cite{lobo2008auc}: The Area under the ROC curve, referred to as AUC, is the most widely used metric for evaluating saliency maps. The saliency map is treated as a binary classifier of fixations at various threshold values (level sets), and a ROC curve is swept out by measuring the true and false positive rates under each binary classifier (level set).

\item [$\bm{-}$] \textbf{aIOU }\cite{rahman2016optimizing}: IoU is the most commonly used metric for comparing the similarity between two arbitrary shapes. The IoU measure gives the similarity between the predicted region and the ground-truth region, and is defined as the size of the intersection divided by the union of the two regions. It can be formulated as:

\begin{equation}
    IoU = \frac{TP}{TP+FP+FN},
\end{equation}
where $TP$, $FP$, and $FN$ denote the true positive, false positive, and false negative counts, respectively.

\item [$\bm{-}$] \textbf{SIM }\cite{swain1991color}: The similarity metric (SIM) measures the similarity between the prediction map and the ground truth map. Given a prediction map $P$ and a continuous ground truth map $Q^{D}$, $SIM(\cdot)$ is computed as the sum of the minimum values at each element, after normalizing the input maps:
\begin{equation}
\begin{split}
\small
   &SIM (P,Q^{D})=\sum_{i}min(P_{i},Q_{i}^{D}),\\
    & where\quad \sum_{i}P_{i}=\sum_{i}Q_{i}^{D}=1. \label{eq:no21}
\end{split}
\end{equation}

\item [$\bm{-}$] \textbf{MAE }\cite{willmott2005advantages}: The Mean Absolute Error (MAE) is a useful measure widely used in model evaluations. The calculation of MAE is relatively simple. It involves summing the magnitudes (absolute values) of the errors to obtain the ``total error'' and then dividing the total error by $n$:

\begin{equation}
\mathrm{MAE}=\frac{1}{n} \sum_{i=1}^n\left|e_i\right|,
\end{equation}
where $e_i$ is the calculated model error.
\end{itemize}

\subsection*{A.3. Training Details}
\addcontentsline{toc}{subsection}{\textcolor[rgb]{0,0,0}{A.3. Training Details}}
Our model is implemented in PyTorch and trained with the Adam optimizer. The training epoch is set to 80. All training processes are on a single NVIDIA 3090 Ti GPU with an initial learning rate of 0.0001. The loss balance hyper-parameters  $\lambda_1$, $\lambda_2$, $\lambda_3$ are set to 1, 0.3, and 0.5 respectively, and the training batch size is set to 16. The image extractor uses the pre-trained parameters on ImageNet, while the point cloud extractor is trained from scratch. In addition, since images and point clouds do not need strict one-to-one pairing, we pair images and point clouds online during training. An image could be paired with $n$ point clouds in one training step, which is equivalent to expanding training samples. In the training process, the loss of an image and all paired point clouds is accumulated, and the gradient is calculated according to the accumulated loss. We set $n=2$ in our implementation.

\section*{B. Dataset}
\addcontentsline{toc}{section}{\textcolor[rgb]{0,0,0}{B. Dataset}}
\label{Dataset}
\subsection*{B.1. Data Samples}
\begin{figure}[t]
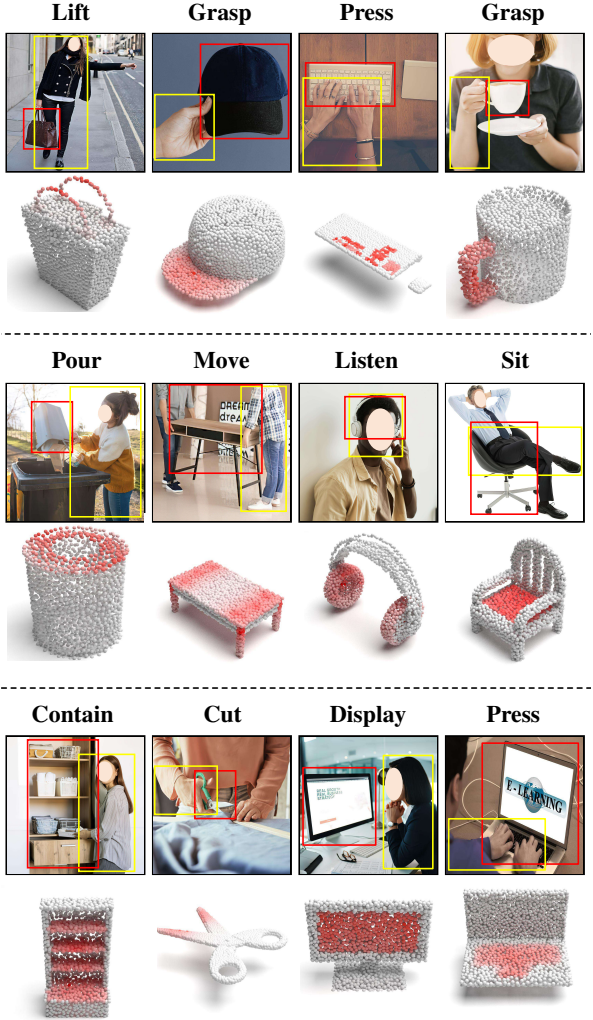

	\centering
        \small
		\begin{overpic}[width=0.94\linewidth]{figs/sup_dataset.pdf}
            \put(5,65){\textbf{Pour}}
            \put(19,65){\textbf{Move}}
            \put(33,65){\textbf{Listen}}
             \put(49.5,65){\textbf{Sit}}
             
            \put(3,30){\textbf{Contain}}        
            \put(20,30){\textbf{Cut}}
            \put(32.5,30){\textbf{Display}}
            \put(48,30){\textbf{Press}}
            
            \put(5,99.5){\textbf{Lift}}
            \put(18.5,99.5){\textbf{Grasp}}
            \put(33.5,99.5){\textbf{Press}}
            \put(47.5,99.5){\textbf{Grasp}}
	\end{overpic}
	\caption{\textbf{Examples of PIAD.} Some paired images and point clouds in PIAD. The ``yellow'' box in the image is the bounding box of the interactive subject, and the ``red'' box is the bounding box of the interactive object.}
	\label{Datasetexample}
\end{figure}

\addcontentsline{toc}{subsection}{\textcolor[rgb]{0,0,0}{A.1. Data samples}}
According to our collection principles, the object in the image and point cloud belongs to the same category and the image demonstrates the way in which the 3D object could interact. For the interactive subject in the image, it could be a human or just a human body part, this is in line with the way humans learn from demonstrations, which usually only needs to observe the specific parts that occur interactions. Here, we give more data pairs in \textbf{P}oint-Image \textbf{A}ffordance \textbf{D}ataset (\textbf{PIAD}), shown in Fig. \ref{Datasetexample}.

\begin{table*}[]
  \centering
  \renewcommand{\arraystretch}{1.}
  \renewcommand{\tabcolsep}{15pt}
   \caption{\textbf{Data Distribution.} The affordance corresponding to each object and the number of its corresponding images and point clouds.}
   \label{dataset:Seen}
   \vspace{5pt}
\begin{tabular}{c|c|cc|cc}
\toprule
\multirow{2}{*}{\textbf{Objects}} & \multirow{2}{*}{\textbf{Affordance}} & \multicolumn{2}{c|}{\textbf{Image}}                   & \multicolumn{2}{c}{\textbf{Point}} \\ \cmidrule{3-6} 
                                  &                                      & \textbf{Train} & \multicolumn{1}{c|}{\textbf{Test}} & \textbf{Train}   & \textbf{Test}   \\ \midrule
\textbf{Vase}                     & wrapgrasp,contain                    & 186            & 46                                 & 209              & 46              \\
\textbf{Display}                  & display                              & 210            & 52                                 & 253              & 52              \\
\textbf{Bed}                      & lay,sit                              & 117            & 28                                 & 127              & 28              \\
\textbf{Microwave}                & contain,open                         & 111            & 27                                 & 130              & 27              \\
\textbf{Door}                     & push,open                            & 105            & 25                                 & 108              & 25              \\
\textbf{Earphone}                 & listen,grasp                         & 144            & 35                                 & 157              & 35              \\
\textbf{Bottle}                   & wrapgrasp,contain,open,pour          & 252            & 61                                 & 288              & 61              \\
\textbf{Bowl}                     & wrapgrasp,contain,pour               & 117            & 28                                 & 132              & 28              \\
\textbf{Laptop}                   & display,press                        & 238            & 58                                 & 295              & 58              \\
\textbf{Clock}                    & display                              & 38             & 9                                  & 205              & 9               \\
\textbf{Scissors}                 & cut,stab,grasp                       & 43             & 10                                 & 49               & 10              \\
\textbf{Mug}                      & wrapgrasp,contain,grasp,pour         & 142            & 35                                 & 152              & 35              \\
\textbf{Faucet}                   & open,grasp                           & 196            & 48                                 & 210              & 48              \\
\textbf{StorageFurniture}         & contain,open                         & 217            & 53                                 & 329              & 53              \\
\textbf{Bag}                      & contain,open,grasp,lift              & 74             & 15                                 & 88               & 15              \\
\textbf{Chair}                    & sit,move                             & 797            & 199                                & 1352             & 199             \\
\textbf{Dishwasher}               & contain,open                         & 84             & 20                                 & 117              & 20              \\
\textbf{Refrigerator}             & contain,open                         & 119            & 28                                 & 130              & 28              \\
\textbf{Table}                    & move,support                         & 401            & 100                                & 957              & 100             \\
\textbf{Hat}                      & wear,grasp                           & 143            & 35                                 & 156              & 35              \\
\textbf{Keyboard}                 & press                                & 100            & 25                                 & 110              & 25              \\
\textbf{Knife}                    & cut,stab,grasp                       & 196            & 47                                 & 225              & 47              \\
\textbf{TrashCan}                 & contain,open,pour                    & 120            & 28                                 & 221              & 28              \\ \bottomrule
\end{tabular}
\end{table*}

\subsection*{B.2. Dataset Partitions}
\addcontentsline{toc}{subsection}{\textcolor[rgb]{0,0,0}{B.2. Dataset Partitions}}
Here, we explain how the dataset is divided and the reason for doing so. PIAD includes 23 object categories and 17 affordance classes, we divide it into two partitions: \textbf{\texttt{Seen}} and \textbf{\texttt{Unseen}}. \textbf{\texttt{Seen}} includes all objects and affordances, this partition is utilized to verify whether it is feasible to ground 3D object affordance in such a learning paradigm. The category of affordance corresponding to each object and the number of images and point clouds in the training set and testing set are shown in Tab. \ref{dataset:Seen}. A series of experiments in the main paper has proved this kind of learning paradigm is achievable. However, embodied agents commonly face the human space, and meet novel objects, to facilitate the agents' capabilities for anticipating novel objects' affordance, we make the \textbf{\texttt{Unseen}} partition. In \textbf{\texttt{Unseen}}, serval objects do not exist in the training set. The principle of the partition is that the affordance of unseen objects should have corresponding objects in the training set. Eventually, we chose the following objects to the testing set as the unseen category: ``Dishwasher'', ``Microwave'', ``Scissors'', ``Laptop'', ``Bed'', ``Vase''. It can be seen from Tab. \ref{dataset:Seen} that the number of images and point clouds is not consistent. They do not need a fixed one-to-one pair, an image can be paired with multiple point clouds.

\section*{C. Experiments}
\addcontentsline{toc}{section}{\textcolor[rgb]{0,0,0}{C. Experiments}}

\begin{table*}[h]
\centering
\small
  \renewcommand{\arraystretch}{1.}
  \renewcommand{\tabcolsep}{1pt}
  \caption{{\textbf{Evaluation Metrics in \textbf{\texttt{Seen}}.} Objective results of each affordance type for all comparison methods in the \textbf{\texttt{Seen}}. ``cont.'' denotes ``contain'', ``supp.'' denotes ``support'', ``wrap.'' denotes ``wrapgrasp'', and ``disp.'' denotes ``display''.}}
\label{table:results_Seen}
\vspace{5pt}
\begin{tabular}{c|c|ccccccccccccccccc}
\toprule
\textbf{Method}                   & \textbf{Metrics}            & \textbf{grasp} & \textbf{cont.} & \textbf{lift} & \textbf{open} & \textbf{lay} & \textbf{sit} & \textbf{supp.} & \textbf{wrap.} & \textbf{pour} & \textbf{move} & \textbf{disp.} & \textbf{push} & \textbf{listen} & \textbf{wear} & \textbf{press} & \textbf{cut} & \textbf{stab} \\ \midrule
\multirow{4}{*}{\textbf{\texttt{Baseline}}}    & \textbf{AUC}     & 63.31          & 51.94            & 91.1          & 75.17         & 59.84        & 76.43        & 71.24            & 52.83              & 83.79         & 60.23         & 73.01            & 62.16         & 51.20            & 53.99         & 72.82          & 80.46        & 61.13         \\
                                  & \textbf{aIOU}     & 3.89           & 4.65             & 9.55          & 4.17          & 6.69         & 8.70         & 7.33             & 3.88               & 7.04          & 4.69          & 7.66             & 4.20          & 5.21            & 4.16          & 4.93           & 5.32         & 5.57          \\
                                  & \textbf{SIM}     & 0.387          & 0.352            & 0.155         & 0.146         & 0.305        & 0.363        & 0.524            & 0.577              & 0.418         & 0.387         & 0.288            & 0.557         & 0.356           & 0.539         & 0.102          & 0.478        & 0.209         \\
                                  & \textbf{MAE}     & 0.157          & 0.154            & 0.129         & 0.121         & 0.214        & 0.177        & 0.164            & 0.145              & 0.125         & 0.158         & 0.199            & 0.089         & 0.192           & 0.154         & 0.140          & 0.129        & 0.137         \\ \midrule
\multirow{4}{*}{\textcolor[rgb]{0.2,0.8,0.1}{\texttt{\textbf{MBDF}}} \cite{tan2021mbdf}}    & \textbf{AUC}     & 58.18          & 76.21            & 76.70         & 69.35         & 86.71        & 94.72        & 84.84            & 58.00              & 73.60         & 50.25         & 78.99            & 63.73         & 68.09           & 65.60         & 87.25          & 80.99        & 64.75         \\
                                  & \textbf{aIOU}     & 6.10           & 8.60             & 12.06         & 5.20          & 10.79        & 24.11        & 10.51            & 3.97               & 8.07          & 5.83          & 10.70            & 5.96          & 4.19            & 4.49          & 6.59           & 5.63         & 5.87          \\
                                  & \textbf{SIM}    & 0.397          & 0.386            & 0.162         & 0.154         & 0.466        & 0.561        & 0.631            & 0.590              & 0.392         & 0.389         & 0.449            & 0.553         & 0.405           & 0.561         & 0.222          & 0.430        & 0.306         \\
                                  & \textbf{MAE}      & 0.162          & 0.150            & 0.129         & 0.139         & 0.151        & 0.109        & 0.128            & 0.147              & 0.148         & 0.191         & 0.147            & 0.121         & 0.166           & 0.144         & 0.131          & 0.126        & 0.142         \\ \midrule
\multirow{4}{*}{\textcolor[rgb]{0.2,0.8,0.1}{\texttt{\textbf{PMF}}} \cite{zhuang2021perception}}     & \textbf{AUC}     & 60.90          & 73.90            & 78.03         & 70.73         & 87.48        & 95.22        & 82.66            & 57.56              & 82.71         & 50.14         & 80.22            & 63.94         & 61.34           & 61.53         & 87.96          & 82.09        & 65.44         \\
                                  & \textbf{aIOU}     & 7.12           & 8.41             & 11.49         & 6.94          & 14.52        & 25.50        & 7.83             & 4.21               & 8.92          & 5.91          & 15.86            & 6.18          & 3.15            & 2.87          & 6.76           & 3.03         & 4.09          \\
                                  & \textbf{SIM}     & 0.426          & 0.385            & 0.193         & 0.180         & 0.486        & 0.598        & 0.653            & 0.565              & 0.397         & 0.380         & 0.470            & 0.535         & 0.412           & 0.548         & 0.237          & 0.421        & 0.356         \\
                                  & \textbf{MAE}     & 0.161          & 0.148            & 0.164         & 0.143         & 0.143        & 0.098        & 0.126            & 0.152              & 0.147         & 0.186         & 0.137            & 0.115         & 0.122           & 0.154         & 0.121          & 0.132        & 0.148         \\ \midrule
\multirow{4}{*}{\textcolor[rgb]{0.2,0.8,0.1}{\texttt{\textbf{FRCNN}}} \cite{xu2022fusionrcnn}}   & \textbf{AUC}     & 62.56          & 77.88            & 77.71         & 74.41         & 88.69        & 95.72        & 82.59            & 53.73              & 77.38         & 53.73         & 79.53            & 62.51         & 69.55           & 68.83         & 87.46          & 80.24        & 68.84         \\
                                  & \textbf{aIOU}     & 7.36           & 9.12             & 11.51         & 7.39          & 15.03        & 25.17        & 7.76             & 4.05               & 7.05          & 6.29          & 15.93            & 5.17          & 4.70            & 2.79          & 6.58           & 4.97         & 4.41          \\
                                  & \textbf{SIM}     & 0.423          & 0.387            & 0.182         & 0.189         & 0.494        & 0.591        & 0.644            & 0.559              & 0.399         & 0.376         & 0.475            & 0.543         & 0.431           & 0.560         & 0.234          & 0.421        & 0.385         \\
                                  & \textbf{MAE}     & 0.148          & 0.144            & 0.160         & 0.137         & 0.137        & 0.094        & 0.124            & 0.132              & 0.136         & 0.173         & 0.136            & 0.116         & 0.162           & 0.148         & 0.119          & 0.131        & 0.132         \\ \midrule
\multirow{4}{*}{\textcolor[rgb]{0.99,0.5,0.0}{\textbf{\texttt{ILN}}} \cite{chen2022imlovenet}}     & \textbf{AUC}     & 62.45          & 76.43            & 77.17         & 74.84         & 88.08        & 94.99        & 82.77            & 51.84              & 82.59         & 52.15         & 80.08            & 62.93         & 68.68           & 64.34         & 87.84          & 82.59        & 66.51         \\
                                  & \textbf{aIOU}     & 8.02           & 8.13             & 10.42         & 6.69          & 16.35        & 26.72        & 7.90             & 4.97               & 8.18          & 5.97          & 17.25            & 4.84          & 5.10            & 3.46          & 6.93           & 5.51         & 4.22          \\
                                  & \textbf{SIM}     & 0.393          & 0.388            & 0.212         & 0.183         & 0.512        & 0.613        & 0.654            & 0.557              & 0.394         & 0.386         & 0.497            & 0.547         & 0.410           & 0.548         & 0.231          & 0.440        & 0.379         \\
                                  & \textbf{MAE}     & 0.144          & 0.148            & 0.158         & 0.141         & 0.133        & 0.095        & 0.127            & 0.136              & 0.147         & 0.172         & 0.132            & 0.107         & 0.173           & 0.146         & 0.108          & 0.123        & 0.139         \\ \midrule
\multirow{4}{*}{\textcolor[rgb]{0.99,0.5,0.0}{\textbf{\texttt{PFusion}}} \cite{xu2018pointfusion}} & \textbf{AUC}     & 65.12          & 78.43            & 84.52         & 74.12         & 89.50        & 95.47        & 82.88            & 51.00              & 82.02         & 52.37         & 84.69            & 67.45         & 69.98           & 67.08         & 87.63          & 85.94        & 67.30         \\
                                  & \textbf{aIOU}     & 9.25           & 9.86             & 11.43         & 8.48          & 12.07        & 25.39        & 7.59             & 4.95               & 6.07          & 6.22          & 18.14            & 3.83          & 6.63            & 4.68          & 7.42           & 6.22         & 4.26          \\
                                  & \textbf{SIM}     & 0.387          & 0.392            & 0.223         & 0.156         & 0.445        & 0.616        & 0.651            & 0.573              & 0.396         & 0.387         & 0.505            & 0.569         & 0.490           & 0.554         & 0.230          & 0.402        & 0.385         \\
                                  & \textbf{MAE}     & 0.139          & 0.138            & 0.133         & 0.132         & 0.138        & 0.095        & 0.125            & 0.131              & 0.157         & 0.196         & 0.128            & 0.096         & 0.173           & 0.148         & 0.103          & 0.142        & 0.155         \\ \midrule
\multirow{4}{*}{\textcolor[rgb]{0.99,0.5,0.0}{\textbf{\texttt{XMF}}} \cite{aiello2022cross}}     & \textbf{AUC}     & 62.98          & 79.97            & 78.93         & 79.74         & 88.93        & 94.91        & 84.56            & 56.47              & 82.15         & 54.97         & 82.91            & 63.74         & 74.93           & 70.57         & 88.18          & 79.17        & 69.97         \\
                                  & \textbf{aIOU}     & 12.82          & 10.41            & 13.24         & 9.17          & 17.00        & 26.65        & 7.89             & 5.24               & 7.81          & 6.46          & 17.25            & 2.68          & 5.31            & 5.34          & 7.87           & 6.35         & 5.88          \\
                                  & \textbf{SIM}     & 0.415          & 0.401            & 0.334         & 0.184         & 0.427        & 0.619        & 0.659            & 0.569              & 0.399         & 0.391         & 0.508            & 0.535         & 0.433           & 0.565         & 0.237          & 0.427        & 0.394         \\
                                  & \textbf{MAE}     & 0.121          & 0.131            & 0.138         & 0.131         & 0.129        & 0.093        & 0.119            & 0.128              & 0.159         & 0.192         & 0.127            & 0.083         & 0.164           & 0.129         & 0.113          & 0.135        & 0.152         \\ \midrule
\multirow{4}{*}{\textbf{\texttt{Ours}}}    & \textbf{AUC}     & 77.53          & 83.84            & 95.05         & 90.89         & 93.54        & 95.94        & 84.58            & 66.71              & 86.02         & 63.09         & 89.29            & 84.71         & 87.13           & 71.16         & 89.46          & 86.69        & 76.41         \\
                                  & \textbf{aIOU}     & 16.83          & 17.12            & 31.95         & 28.39         & 31.80        & 37.72        & 12.04            & 6.02               & 20.33         & 5.57         & 30.57            & 1.79          & 15.59           & 6.55          & 14.42          & 12.95        & 9.48         \\
                                  & \textbf{SIM}    & 0.530          & 0.534            & 0.368         & 0.401         & 0.685         & 0.723        & 0.716            & 0.571              & 0.525         & 0.443         & 0.657            & 0.418         & 0.671           & 0.563         & 0.402          & 0.507        & 0.280         \\
                                  & \textbf{MAE}     & 0.108          & 0.093            & 0.030         & 0.044         & 0.081        & 0.066        & 0.100            & 0.143              & 0.096         & 0.174         & 0.084            & 0.085         & 0.090           & 0.129         & 0.059          & 0.085        & 0.099         \\ \bottomrule
\end{tabular}
\end{table*}

\subsection*{C.1. Details of Modular Baselines}
\addcontentsline{toc}{subsection}{\textcolor[rgb]{0,0,0}{C.1. Details of Modular Baselines}}
\par Since there is no previous work research grounding 3D object affordance in such a cross-modal fashion, we select some advanced studies in the image-point cloud cross-modal learning area to make the comparison experiment. The common aspect of these works is that they extract the feature of images and point clouds separately, and then align or fuse the extracted features. These methods are divided into two types: one is to use the camera intrinsic parameter in the alignment or fusion module for obtaining spatial correspondence, and the other is to align or fusion multi-modal features directly in the feature space without using the intrinsics. For methods using the camera intrinsic parameter, we remove the step that uses intrinsic parameters to verify the effectiveness of such methods on the proposed task setting, in which images and point clouds originate from different physical instances and it is infeasible to obtain spatial correspondence through the assistance of camera priors. All comparative methods share the same extractors with our method, and the difference occurs subsequent to the extraction of $\mathbf{F}_{p}$ and $\mathbf{F}_{i}$.
\begin{itemize}

\item [$\bm{-}$]\texttt{\textbf{Baseline}}: For the design of baseline, we directly concatenate the features that are output from the image and point cloud extractors, let the concatenation be a fusion block and there are no intermediate steps to correspond the region features from different sources.

\item [$\bm{-}$] \textcolor[rgb]{0.2,0.8,0.1}{\texttt{\textbf{MBDF-Net (MBDF)}}} \cite{tan2021mbdf}: This work focus on 3D object detection work. It has three branches: image branch, lidar branch, and fusion branch. An Adaptive Attention Fusion Module (AAF) is proposed in this work to fuse the features from the image and point cloud. We take its AAF as the cross-modal block to fuse the image and point cloud feature. And in AAF, we remove the step that utilizes the inner parameters of the camera to project the coordinates $p(x,y,z)$ of each point in the point cloud into the corresponding image coordinate $p{'}({x^{'},y^{'}})$.

\item [$\bm{-}$] \textcolor[rgb]{0.2,0.8,0.1}{\textbf{\texttt{PMF}}} \cite{zhuang2021perception}: For 3D point cloud segmentation, this paper designs a two-stream network to fuse the rich semantic information provided by images with point cloud features to obtain more fine-grained results. For the multi-modal feature fusion, it devises a residual-based fusion model to concatenate image and point cloud features and uses convolution and attention to calculate the fusion feature, and finally, residual connection with the point cloud features. We use the RF module to fuse features that are from different sources and remove the step which projects the point cloud to the camera coordinate system by perspective projection.

\item [$\bm{-}$] \textcolor[rgb]{0.2,0.8,0.1}{\textbf{\texttt{FusionRCNN (FRCNN)}}} \cite{xu2022fusionrcnn}: This work first extracts proposals in the image and point cloud respectively and then fuses these proposals by cross-attention and self-attention. Specifically, for the extracted point cloud and image features, performing a self-attention on both features, then, the point cloud feature is regarded as a query, and the image feature is regarded as key and value, using a cross-attention to fuse them. In the whole pipeline, this operation is performed twice. We fuse the image and point cloud feature with the aforementioned scheme.

\begin{table*}[]
\centering
  \renewcommand{\arraystretch}{1.}
  \renewcommand{\tabcolsep}{1pt}
  \caption{{\textbf{Evaluation Metrics in \textbf{\texttt{Unseen}}.} Objective results of each affordance type for all comparison methods in the \textbf{\texttt{Useen}}. ``cont.'' denotes ``contain'', ``wrap.'' denotes ``wrapgrasp''}}
\label{table:obj_useen}
\vspace{3pt}
\begin{tabular}{c|c|cccccccccc}
\toprule
\textbf{Method}                      & \textbf{Metrics}       & \textbf{cont.}           & \textbf{lay} & \textbf{sit} & \textbf{wrap.} & \textbf{open} & \textbf{display} & \textbf{stab} & \textbf{grasp} & \textbf{press} & \textbf{cut} \\ \midrule
\multirow{4}{*}{\textbf{\texttt{Baseline}}} & \textbf{AUC}  & \multicolumn{1}{c}{60.15} & 76.08        & 63.89        & 40.15          & 72.41         & 37.84            & 51.37         & 43.77          & 61.51          & 69.55        \\
                            & \textbf{aIOU} & 4.49                       & 10.31        & 5.16         & 1.30           & 3.30          & 2.88             & 3.06          & 2.41           & 4.18           & 5.39         \\
                            & \textbf{SIM}  & 0.351                      & 0.450        & 0.370        & 0.448          & 0.126         & 0.061            & 0.267         & 0.166          & 0.228          & 0.379        \\
                            & \textbf{MAE}  & 0.157                      & 0.153        & 0.157        & 0.181          & 0.117         & 0.385            & 0.147         & 0.146          & 0.141          & 0.112        \\ \midrule 
\multirow{4}{*}{\textcolor[rgb]{0.2,0.8,0.1}{\texttt{\textbf{MBDF}}} \cite{tan2021mbdf}} & \textbf{AUC}  & 62.81                      & 76.80        & 64.30        & 41.47          & 73.25         & 45.26            & 61.75         & 46.62          & 65.57          & 75.69        \\
                            & \textbf{aIOU} & 5.09                       & 11.28        & 5.63         & 1.52           & 3.68          & 3.94             & 3.35          & 2.42           & 4.64           & 6.36         \\
                            & \textbf{SIM}  & 0.364                      & 0.464        & 0.379        & 0.418          & 0.133         & 0.148            & 0.279         & 0.168          & 0.236          & 0.382        \\
                            & \textbf{MAE}  & 0.151                      & 0.149        & 0.152        & 0.177          & 0.108         & 0.286            & 0.124         & 0.150          & 0.137          & 0.098        \\\midrule
\multirow{4}{*}{\textcolor[rgb]{0.2,0.8,0.1}{\texttt{\textbf{PMF}}} \cite{zhuang2021perception}} & \textbf{AUC}  & 64.10                      & 80.54        & 64.89        & 42.02          & 74.86         & 51.62            & 68.93         & 48.44          & 65.98          & 79.05        \\
                            & \textbf{aIOU} & 5.15                       & 13.16        & 5.83         & 1.59           & 3.85          & 3.19             & 4.31          & 2.83           & 4.97           & 7.93         \\
                            & \textbf{SIM}  & 0.368                      & 0.465        & 0.381        & 0.448          & 0.139         & 0.123            & 0.313         & 0.174          & 0.245          & 0.387        \\
                            & \textbf{MAE}  & 0.148                      & 0.147        & 0.153        & 0.172          & 0.102         & 0.262            & 0.113         & 0.144          & 0.135          & 0.094        \\\midrule
\multirow{4}{*}{\textcolor[rgb]{0.2,0.8,0.1}{\texttt{\textbf{FRCNN}}} \cite{xu2022fusionrcnn}} & \textbf{AUC}  & 64.11                      & 84.18        & 66.37        & 44.27          & 74.77         & 48.12            & 71.58         & 49.32          & 67.58          & 82.46        \\
                            & \textbf{aIOU} & 5.54                       & 13.72        & 6.28         & 1.56           & 3.96          & 4.64             & 4.47          & 3.13           & 5.13           & 8.74         \\
                            & \textbf{SIM}  & 0.384                      & 0.481        & 0.389        & 0.463          & 0.143         & 0.131            & 0.341         & 0.189          & 0.263          & 0.394        \\
                            & \textbf{MAE}  & 0.142                      & 0.142        & 0.147        & 0.173          & 0.099         & 0.258            & 0.107         & 0.137          & 0.130          & 0.085        \\\midrule
\multirow{4}{*}{\textcolor[rgb]{0.99,0.5,0.0}{\textbf{\texttt{ILN}}} \cite{chen2022imlovenet}} & \textbf{AUC}  & 66.76                      & 83.17        & 65.87        & 42.21          & 73.75         & 54.5             & 68.48         & 48.97          & 66.51          & 81.42        \\
                            & \textbf{aIOU} & 5.87                       & 13.39        & 5.78         & 1.71           & 3.87          & 4.73             & 4.39          & 3.05           & 5.05           & 8.17         \\
                            & \textbf{SIM}  & 0.382                      & 0.474        & 0.385        & 0.419          & 0.140         & 0.118            & 0.312         & 0.187          & 0.259          & 0.392        \\
                            & \textbf{MAE}  & 0.145                      & 0.145        & 0.151        & 0.167          & 0.096         & 0.274            & 0.112         & 0.129          & 0.132          & 0.091        \\\midrule
\multirow{4}{*}{\textcolor[rgb]{0.99,0.5,0.0}{\textbf{\texttt{PFusion}}} \cite{xu2018pointfusion}} & \textbf{AUC}  & 65.56                      & 83.35        & 67.54        & 42.70          & 76.03         & 56.93            & 69.05         & 52.23          & 68.81          & 82.39        \\
                            & \textbf{aIOU} & 5.92                       & 13.56        & 6.83         & 1.63           & 4.14          & 5.14             & 4.27          & 3.79           & 5.25           & 9.64         \\
                            & \textbf{SIM}  & 0.396                      & 0.483        & 0.391        & 0.47           & 0.151         & 0.072            & 0.334         & 0.234          & 0.262          & 0.413        \\
                            & \textbf{MAE}  & 0.140                      & 0.139        & 0.145        & 0.163          & 0.096         & 0.255            & 0.091         & 0.123          & 0.128          & 0.083        \\ \midrule
\multirow{4}{*}{\textcolor[rgb]{0.99,0.5,0.0}{\textbf{\texttt{XMF}}} \cite{aiello2022cross}} & \textbf{AUC}  & 67.98                      & 84.02        & 68.45        & 45.74          & 78.53         & 62.2             & 76.92         & 59.19          & 69.32          & 85.87        \\
                            & \textbf{aIOU} & 6.29                       & 15.10        & 7.29         & 1.42           & 4.32          & 6.20             & 6.12          & 3.97           & 5.71           & 13.95        \\
                            & \textbf{SIM}  & 0.412                      & 0.503        & 0.403        & 0.451          & 0.156         & 0.075            & 0.351         & 0.278          & 0.270          & 0.435        \\
                            & \textbf{MAE}  & 0.137                      & 0.135        & 0.144        & 0.156          & 0.094         & 0.240            & 0.087         & 0.117          & 0.124          & 0.078        \\ \midrule
\multirow{4}{*}{\textbf{\texttt{Ours}}} & \textbf{AUC}  & 67.96                     & 84.82        & 71.10        & 56.39          & 90.91         & 85.51            & 98.83         & 78.60          & 68.07          & 95.95        \\
                            & \textbf{aIOU} & 7.24                       & 18.12        & 8.47         & 1.89           & 12.28          & 16.28            & 10.39         & 4.79           & 4.22           & 21.47        \\
                            & \textbf{SIM}  & 0.430                      & 0.525        & 0.407        & 0.556          & 0.227         & 0.393            & 0.437         & 0.533          & 0.194          & 0.599        \\
                            & \textbf{MAE}  & 0.125                     & 0.130        & 0.143        & 0.150          & 0.050         & 0.130            & 0.044         & 0.102          & 0.122          & 0.057        \\ \bottomrule
\end{tabular}
\end{table*}

\item [$\bm{-}$] \textcolor[rgb]{0.99,0.5,0.0}{\textbf{\texttt{ImloveNet (ILN)}}} \cite{chen2022imlovenet}: This research uses images to support the registration of low-overlap point cloud pairs. Its purpose is to use images to provide information for the low-overlap regions of point cloud pairs, so as to support the registration, it is the SOTA on low-overlap point cloud pairs registration task. It also extracts the image and point cloud features separately, and in the fusion module, it projects image features into the 3D feature space through a learnable mapping. Then, applying the attention mechanism fuses point cloud feature, image feature, and the projected 3D feature in turn. We take this mechanism to fuse image and point cloud features in implementation.

\item [$\bm{-}$] \textcolor[rgb]{0.99,0.5,0.0}{\textbf{\texttt{PointFusion (PFusion)}}} \cite{xu2018pointfusion}: This is an early work towards 3D object detection. It also extracts the features of the point cloud and image respectively. For the fusion of different modal features, its processing method is relatively simple. The image branch eventually outputs a global feature, while the point cloud branch outputs a global feature and point-wise feature, the two global features and the point-wise feature do dense fusion to get the fusion feature finally. And we apply this operation to implement the fusion of image and point cloud features.

\item [$\bm{-}$] \textcolor[rgb]{0.99,0.5,0.0}{\textbf{\texttt{XMFnet (XMF)}}} \cite{aiello2022cross}: This study focus on point cloud completion, it is the SOTA on cross-modal point cloud completion task. It proposes XMFnet, which is composed of two modality-specific feature extractors that capture localized features of the input point cloud and image, then, it uses the combined cross-attention and self-attention to fuse the features of the two modalities. And we apply this block to compute the image and point cloud feature in the pipeline.
\end{itemize}

\subsection*{C.2. Metrics of Each Affordance}
\addcontentsline{toc}{subsection}{\textcolor[rgb]{0,0,0}{C.2. Metrics of Each Affordance}}

\par We give the overall results of each method in the main paper. Here, we display the results of each affordance respectively. The experimental results of all methods in \textbf{\texttt{Seen}} are shown in Tab. \ref{table:results_Seen}. And \textbf{\texttt{Unseen}} results are shown in Tab. \ref{table:obj_useen}. As can be seen, our method achieves the best results under most affordance categories, which demonstrates the superiority of our method in grounding 3D object affordance. These results indicate that our model has great performance whether it is for unseen objects or structures that have not been mapped to a certain affordance, which also indicates the stability and the generalization of our model. At the same time, other methods also achieve considerable objective results under our setting, which proves the rationality of the setting.

\begin{table}[]
\centering
\renewcommand{\arraystretch}{1.}
\renewcommand{\tabcolsep}{4.pt}
   \caption{\textbf{Techniques.}
   Results of different techniques for projecting the region relevance in \textbf{\texttt{Seen}} and \textbf{\texttt{Unseen}}. S-Atten. denotes self-attention, EM-Atten. denotes Expectation-Maximization Attention, and MLP denotes multilayer perception.}
\label{table:projection_method}
\vspace{5pt}
\begin{tabular}{c|c|ccc}
\toprule
\textbf{Setting} & \textbf{Metrics} & \textbf{S-Atten.} & \textbf{EM-Atten.} & \textbf{MLP} \\ 
 \midrule
\multirow{4}{*}{\textbf{\texttt{Seen}}} & \textbf{AUC} &85.16 & 82.37 & 82.93 \\
 &\textbf{aIOU}  & 21.20 & 16.03 & 17.56  \\
 & \textbf{SIM}  & 0.564 & 0.521 & 0.527  \\ 
 & \textbf{MAE}  & 0.088 & 0.102 & 0.948  \\ 
 \midrule
\multirow{4}{*}{\textbf{\texttt{Unseen}}} & \textbf{AUC} & 73.69 & 67.45 & 68.12 \\
 &\textbf{aIOU}  & 8.70 & 6.78 & 7.04  \\
 & \textbf{SIM}  & 0.383 & 0.429 & 0.432  \\ 
 & \textbf{MAE}  & 0.117 & 0.174 & 0.159  \\
 \bottomrule
\end{tabular}
\end{table}

\begin{table}[]
\centering
\renewcommand{\arraystretch}{1.}
\renewcommand{\tabcolsep}{4.pt}
   \caption{{\textbf{Different Backbones.}
   Results of \textbf{\texttt{Seen}} and \textbf{\texttt{Unseen}} settings in different backbone networks. En. P indicates the extractor of the point cloud, En. I indicates the extractor of the image. PN is PointNet++ \cite{qi2017pointnet++}, PM is PointMLP \cite{ma2022rethinking} and Res is ResNet \cite{he2016deep}.}}
\label{table:backbones}
\vspace{5pt}
\begin{tabular}{c|c|c|cccc}
\toprule
\textbf{Setting} & \textbf{En. P} & \textbf{En. I} & \textbf{AUC} & \textbf{aIOU} & \textbf{SIM} & \textbf{MAE} \\ 
 \midrule
\multirow{6}{*}{\textbf{\texttt{Seen}}} & \multirow{3}{*}{\textbf{PN}} &Res18 & 85.16 & 21.20 & 0.564 & 0.088 \\
 & &  Res34  & 85.45 & 21.32 & 0.569 & 0.086 \\
 & &  Res50 & 85.52 & 21.40 & 0.569 & 0.082 \\ 
 \cmidrule{2-7} 
 & \multirow{3}{*}{\textbf{PM}} & Res18 & 84.89 & 19.47 & 0.543 & 0.095 \\
 & & Res34 & 84.98 & 19.66 & 0.548 & 0.088 \\
 & & Res50 & 85.31 & 19.93 & 0.554 & 0.084 \\
 \midrule
\multirow{6}{*}{\textbf{\texttt{Unseen}}} & \multirow{3}{*}{\textbf{PN}} & Res18 & 73.69 & 8.70 & 0.383 & 0.117 \\
 & & Res34 & 73.78 & 8.70 & 0.387 & 0.107 \\
 & & Res50& 73.83 & 8.82 & 0.393 & 0.101 \\ 
 \cmidrule{2-7} 
 & \multirow{3}{*}{\textbf{PM}} & Res18 & 70.11 & 8.29 & 0.436 & 0.146 \\
 & & Res34 & 70.76 & 8.67 & 0.440 & 0.142 \\
 & & Res50 & 71.05 & 8.83 & 0.447 & 0.136 \\
 \bottomrule
\end{tabular}
\end{table}

\subsection*{C.3. Techniques for Establishing Relevance}
\label{sec:projection}
\addcontentsline{toc}{subsection}{\textcolor[rgb]{0,0,0}{C.3. Projection Techniques}}
\par To investigate the way of mapping the region relevance, we conduct a comparative experiment to explore the performance of the aforementioned three techniques. They are transformer-based (self-attention),  expectation-maximization attention (EMA), and multilayer perception (MLP). The results of metrics in \textbf{\texttt{Seen}} and \textbf{\texttt{Unseen}} are shown in Tab. \ref{table:projection_method}. As can be seen from the table, MLP and EMA get sub-optimal performance. We analyze the possible reason is that the region features are derived from different sources, so there are gaps among corresponding region features, MLP cannot effectively establish the mapping. Similarly, EMA is also difficult to excavate the region correlation with a group of bases. While self-attention calculates the correlation between every two regions from different sources, there exists a certain relative difference in these correlations, and this relative difference could make it match the corresponding regions of different sources. Based on the above results, we finally choose the self-attention.

\subsection*{C.4. Different Backbones}
\addcontentsline{toc}{subsection}{\textcolor[rgb]{0,0,0}{C.4. Different Backbones}}

To verify the effectiveness of the framework and the influence of the backbone, we test another backbone network. We use a recently advanced network PointMLP \cite{ma2022rethinking} as the point cloud backbone, it also extracts features of the point cloud hierarchically. In addition, we also test the impact of the model scale on performance, for each point cloud backbone, we take ResNet18, ResNet34 and ResNet50 \cite{he2016deep} as the image feature extractor respectively. The evaluation results are shown in Tab. \ref{table:backbones}, as can be seen from the results, the backbone network does not have a significant impact on the final performance. The larger backbone network could improve the performance, but the improvement is relatively limited. To make the model effective and keep it lightweight, we select PointNet++ and ResNet18 as the final backbone networks.

\subsection*{C.5. Different Hyper-parameters}
\addcontentsline{toc}{subsection}{\textcolor[rgb]{0,0,0}{C.5. Different Hyper-parameters}}
To explore the impact of each hyper-parameter on the total loss, we conduct a comparative experiment of these hyper-parameters. The experimental results are shown in Tab. \ref{table:hyperparameters}. $\lambda_{1}$ is the coefficient of $\mathcal{L}_{HM}$, and it accounts for the highest proportion of the total loss, the reduction of $\lambda_1$ will have a greater impact on the performance. $\lambda_2$ is the coefficient of affordance category loss $\mathcal{L}_{CE}$, and $\lambda_3$ is the coefficient of the KL loss $\mathcal{L}_{KL}$. The best result is to set $\lambda_{2}$ to $0.3$ and $\lambda_{3}$ to 0.5. Whether they increase or decrease, the performance of the model is affected. In addition, we remove $\mathcal{L}_{KL}$ to test the performance, from the result, we can see that lacking $\mathcal{L}_{KL}$ degrades the model performance.

\begin{table}[]
\centering
\small
\renewcommand{\arraystretch}{1.}
\renewcommand{\tabcolsep}{1.5pt}
\caption{\textbf{Hyper-Parameters. }
    The influence of hyper-parameters that balance three losses in the total loss. The last row represents the performance of the model when removing $\mathcal{L}_{KL}$.}
\label{table:hyperparameters}
\vspace{5pt}
\begin{tabular}{ccc|cccc|cccc}
\toprule
\multirow{2}{*}{$\bm{\lambda_{1}}$} & \multirow{2}{*}{$\bm{\lambda_{2}}$} & \multirow{2}{*}{$\bm{\lambda_{3}}$} & \multicolumn{4}{c|}{\textbf{\texttt{Seen}}}                                & \multicolumn{4}{c}{\textbf{\texttt{Unseen}}}\\ 
\cmidrule{4-11} 
                            &                             &                             & \textbf{AUC}   & \textbf{aIOU}  & \textbf{SIM}   & \textbf{MAE}   & \textbf{AUC}   & \textbf{aIOU} & \textbf{SIM}   & \textbf{MAE}   \\ \midrule
1                                               & 0.3                                             & 0.5                                             & \cellcolor{mygray}\textbf{85.16} & \cellcolor{mygray}\textbf{21.20} & \cellcolor{mygray}\textbf{0.564} & \cellcolor{mygray}\textbf{0.088} & \cellcolor{mygray}\textbf{73.69} & \cellcolor{mygray}\textbf{8.70} & \cellcolor{mygray}\textbf{0.383} & \cellcolor{mygray}\textbf{0.117} \\
0.8                                             & 0.3                                             & 0.5                                             & 83.16          & 18.25          & 0.532          & 0.116          & 69.98          & 7.25          & 0.421          & 0.175          \\
1                                               & 0.3                                             & 0.7                                             & 83.78          & 18.93          & 0.537          & 0.112          & 70.13          & 7.52          & 0.429          & 0.168          \\
1                                               & 0.3                                             & 0.3                                             & 83.49          & 18.88          & 0.533          & 0.114          & 70.09          & 7.47          & 0.426          & 0.173          \\
1                                               & 0.5                                             & 0.5                                             & 83.93          & 19.25          & 0.552          & 0.104          & 70.95          & 7.85          & 0.432          & 0.156           \\
1                                               & 0.1                                             & 0.5                                             & 83.42          & 19.12          & 0.546          & 0.115          & 70.17          & 7.60          & 0.425          & 0.163          \\
1                                               & 0.3                                             & 0                                             & 82.42          & 16.94          & 0.528          & 0.183          & 68.23          & 6.92          & 0.402          & 0.273          \\ 
\bottomrule
\end{tabular}
\end{table}

\begin{table}[]
\centering
\renewcommand{\arraystretch}{1.}
\renewcommand{\tabcolsep}{4.pt}
   \caption{\textbf{Pairing Count.} One image could be paired with multiple point clouds for training. Different pairing counts have an influence on the model performance.
   }
\label{table:pairing}
\vspace{5pt}
\begin{tabular}{c|c|cccc}
\toprule
\textbf{Setting} & \textbf{Metrics} & \textbf{1} & \textbf{2} & \textbf{4} & \textbf{6} \\ 
 \midrule
\multirow{4}{*}{\textbf{\texttt{Seen}}} & \textbf{AUC}  &84.75 & 85.16 & 85.44 & 84.82\\
 &\textbf{aIOU}  & 19.45 & 21.20 & 21.83  & 20.15\\
 & \textbf{SIM}  & 0.540 & 0.564 & 0.571  & 0.558\\ 
 & \textbf{MAE}  & 0.095 & 0.088 & 0.083  & 0.090\\ 
 \midrule
\multirow{4}{*}{\textbf{\texttt{Unseen}}} & \textbf{AUC} & 69.98 & 73.69 & 73.72 & 71.32 \\
 &\textbf{aIOU}  & 8.37 & 8.70 & 8.72  & 8.58\\
 & \textbf{SIM}  & 0.375 & 0.383 & 0.391  & 0.380\\ 
 & \textbf{MAE}  & 0.130 & 0.117 & 0.106  & 0.121\\
 \bottomrule
\end{tabular}
\end{table}

\subsection*{C.6. Different Pairing}
\addcontentsline{toc}{subsection}{\textcolor[rgb]{0,0,0}{C.6. Different Pairing}}
Since images and point clouds originate from different physical instances, one image could be paired with multiple point clouds for training, which can increase the diversity of training data pairs and make the trained model more robust. We test the difference in pairing count and the results are in Tab. \ref{table:pairing}. When the number of pairings is set to 2, the model performs well, and when it is set to 4, it achieves better results, but the training time is doubled. If the number of pairs is 6, due to the limitation of computing resources, the batch size has to be reduced, so the performance drops instead. Considering the above situations, we finally chose to set the pairing count to 2 in the implementation.

\subsection*{C.7. More Visual Results}
\addcontentsline{toc}{subsection}{\textcolor[rgb]{0,0,0}{C.7. More Visual Results}}
We show more visual results of comparative methods and our method in \textbf{\texttt{Seen}} and \textbf{\texttt{Unseen}} partitions. Fig. \ref{comparative_seen} shows the result in the \textbf{\texttt{Seen}} and Fig. \ref{comparative_unseen} shows the result in the \textbf{\texttt{Unseen}}. In addition, we also provide more visual results of our method in all partitions, which shows in Fig. \ref{Ours_results}. It can be seen from the visual results that our model is able to anticipate the accurate affordance of object functional components from diverse interactions and across multiple object categories, reflecting its stability, robustness, and generalization. Plus, for methods that directly map affordance to specific structures, we present a visual comparison of results between it and our method, shown in Fig. \ref{fig:paradigm}. It can be seen from the visual results that this type of method is limited in generalization to the unseen structure.

\subsection*{C.8. Partialness and Rotation}
\addcontentsline{toc}{subsection}{\textcolor[rgb]{0,0,0}{C.8. Partialness and Rotation}}
Following the setting proposed by 3D-AffordanceNet \cite{deng20213d}. We make experiments to test the model performance on partial and freely rotating point clouds, which also simulate the occlusion and rotation of objects in the daily environment. The detailed sampling methods of the partial and freely rotating point clouds could be found in 3D-AffordanceNet \cite{deng20213d}. Fig. \ref{fig:rotate} shows the test results. It can be seen from the results that even if the point cloud only has a partial structure or is randomly rotated, our model can still anticipate 3D object affordance on the corresponding geometric structure. This shows the 2D interactions provide various clues for the model to learn the correlation between geometric structure and affordance. And it could be generated for variable object situations.

\section*{D. Potential Applications}
\addcontentsline{toc}{section}{\textcolor[rgb]{0,0,0}{D. Potential Applications}}

\begin{figure}[t]
    \centering
    \footnotesize
    \begin{overpic}[width=0.96\linewidth]{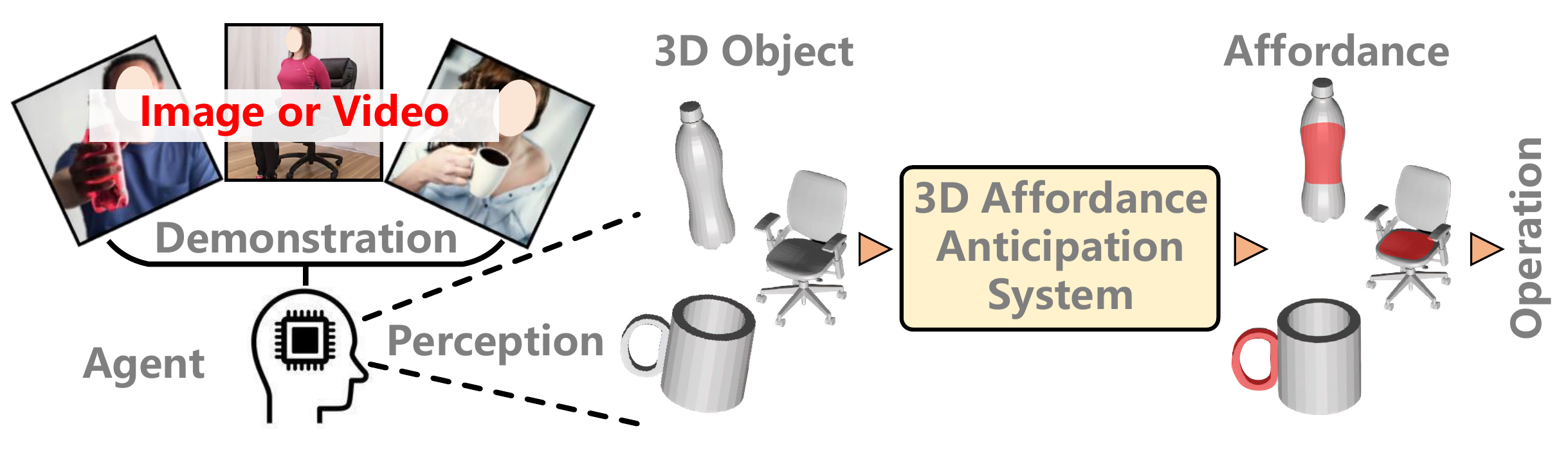}
    \end{overpic}
    \caption{\textbf{Potential Applications.} This work has the potential to bridge the gap between perception and operation, serving areas like demonstration learning \cite{argall2009survey, rahmatizadeh2018virtual}, and may be a part of human-assistant agent system \eg Tesla Bot, Boston Dynamics Atlas \cite{nelson2019petman}.}
    \label{fig:app}
\end{figure}

Object affordance grounding could serve as a link in the embodied system, supporting many down-stream potential applications, as shown in Fig. \ref{fig:app}.

\begin{itemize}

    \item 
    \textbf{Embodied Artificial Intelligence.} 
    Embodied AI \cite{savva2019habitat} is to enable agents to interact with the world from passive perception to active reasoning. The key step to actively understanding the physical world is to know how to interact with the surrounding environment, which is a fundamental skill for embodied agents. And the precondition for the agent to interact with the environment is perceptive the object. Currently, there are some sensors like LIDAR and depth cameras that could sample 3D scene data, and obtain point clouds or depth maps. Meanwhile, many techniques support obtaining a representation of the object from such data. The most direct is to segment or detect the object. Some methods generate the 3D object representation from 2D sources \cite{melas2023realfusion}. The above works ensure obtaining the objects' representation in the scene and support the affordance anticipation system. Anticipating the affordance makes the agent know what action could be done and which location supports the corresponding action on the object representation, which bridges the gap between perception and operation. Such an ability has applications in navigation and manipulation for embodied agents \cite{mandikal2021learning,yang2021collaborative}.

    \item
    \textbf{Imitation Learning.} Imitation learning is a common approach to training intelligent agents to perform tasks by observing demonstrations from humans or other agents. However, a key challenge in imitation learning is the ability to generalize to new scenarios and environments that the agent has not encountered during training. This challenge arises because the agent infers the intentions and goals of the demonstrator from a limited set of observed objects or scenes, and must determine how to adapt those actions to the new context. Affordances are the perceived properties of objects or environments that suggest how they can be used or interacted with. In the context of imitation learning, affordances could help the agent to better understand the demonstrator's intentions and goals by identifying the relevant objects and actions in the environment that the demonstrator is using. By recognizing the affordances, the agent can infer the most likely next action, and can therefore learn to imitate their behaviors more accurately \cite{lopes2007affordance}.
    
    \item 
    \textbf{Augement Reality.} Augmented reality (AR) is currently considered as having potential for daily applications. By anticipating the 3D affordance of objects in the 3D physical world, more practical functions can be brought to AR devices. For example, if an object needs to be repaired, just send a demo image or video to the user, and then the AR device anticipates the corresponding 3D affordance according to the demo to provide operational guidance. It has high application value in such fields as after-sale service, device maintenance, installation industry, and so on. \cite{cheng2013affordances}. 

    \item 
    \textbf{Virtual Reality.} Nowadays, virtual reality (VR) is more and more widely used in the entertainment, online games, and education industries. It provides a virtual environment for people to interact with the three-dimensional virtual scenario. Some online games or entertainment projects will provide some novel interaction scenes to interact with players. An affordance system can play the role of an NPC to provide interaction guidance for users and improve the user experience. \cite{dalgarno2010learning}. 
\end{itemize}

\clearpage

\begin{figure*}[t]
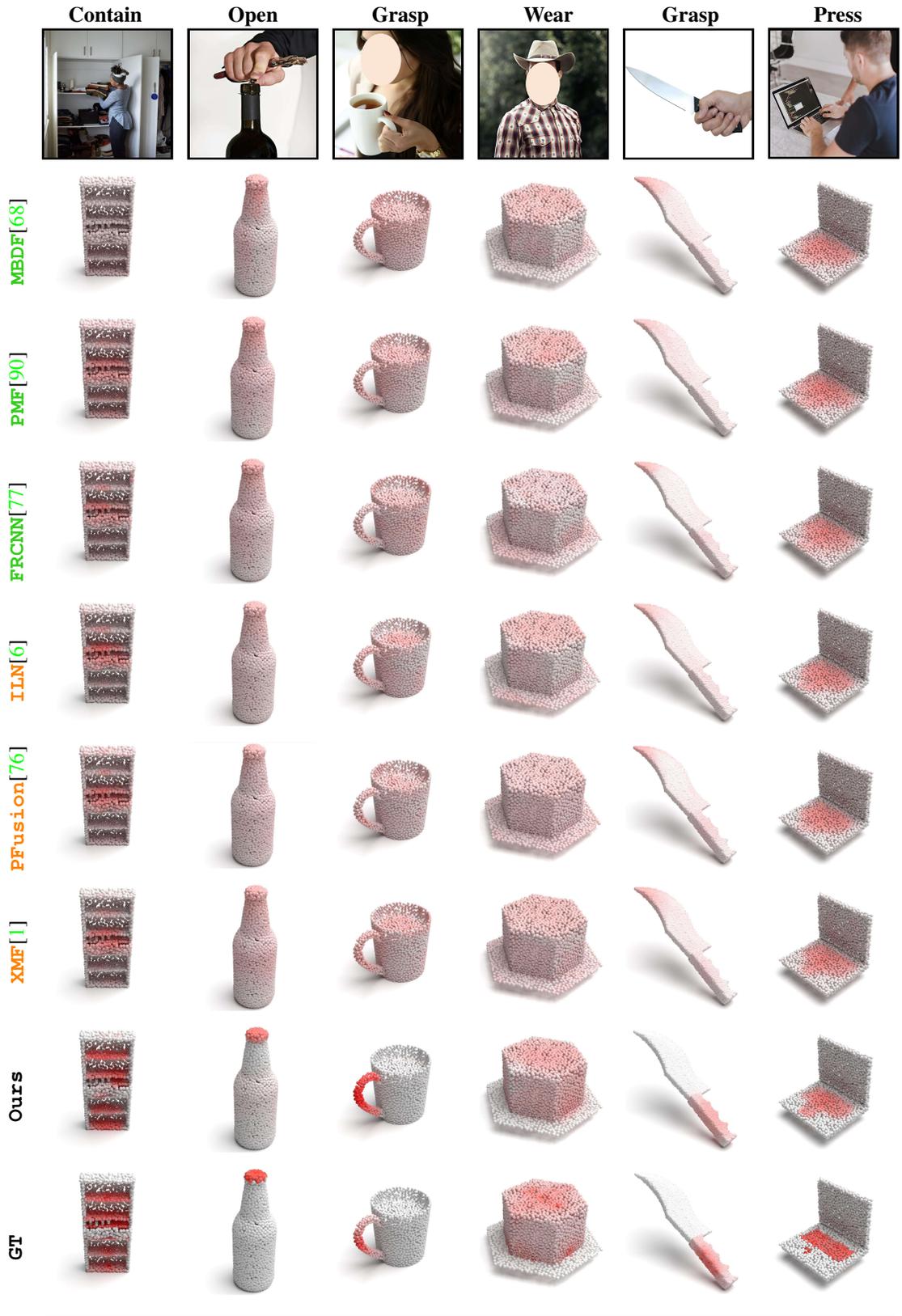

	\centering
		\begin{overpic}[width=0.83\linewidth]{figs/sup_comparative_Seen.pdf}
        \put(3.5,99){\textbf{Contain}}
        \put(15.6,99){\textbf{Open}}
        \put(26.5,99){\textbf{Grasp}}
        \put(38,99){\textbf{Wear}}
        \put(48.5,99){\textbf{Grasp}}
        \put(60,99){\textbf{Press}}
        
        \put(-1,5){\rotatebox{90}{\textbf{\texttt{GT}}}}
        \put(-1,15.5){\rotatebox{90}{\textbf{\texttt{Ours}}}}
        
        \put(-1,26){\rotatebox{90}{\textcolor[rgb]{0.99,0.5,0.0}{\textbf{\texttt{XMF}}}\cite{aiello2022cross}}}
        
        \put(-1,34.5){\rotatebox{90}{\textcolor[rgb]{0.99,0.5,0.0}{\textbf{\texttt{PFusion}}}\cite{xu2018pointfusion}}}
        
        \put(-1,47.3){\rotatebox{90}{\textcolor[rgb]{0.99,0.5,0.0}{\textbf{\texttt{ILN}}}\cite{chen2022imlovenet}}}
        
        \put(-1,56.5){\rotatebox{90}{\textcolor[rgb]{0.2,0.8,0.1}{\textbf{\texttt{FRCNN}}}\cite{xu2022fusionrcnn}}}

        \put(-1,68.3){\rotatebox{90}{\textcolor[rgb]{0.2,0.8,0.1}{\textbf{\texttt{PMF}}}\cite{zhuang2021perception}}}

        \put(-1,79){\rotatebox{90}{\textcolor[rgb]{0.2,0.8,0.1}{\textbf{\texttt{MBDF}}}\cite{tan2021mbdf}}}
	\end{overpic}
	\caption{\textbf{Visual results in \textbf{\texttt{Seen}}.} We give some visual results of all comparison methods and our method in the \textbf{\texttt{Seen}}.}
	\label{comparative_seen}
\end{figure*}

\begin{figure*}[t]
	\centering
		\begin{overpic}[width=0.7\linewidth]{figs/sup_comparative_Unseen.pdf}

        \put(5,99){\textbf{Lay}}
        \put(19.5,99){\textbf{Open}}
        \put(32,99){\textbf{Wrapgrasp}}
        \put(49,99){\textbf{Grasp}}
        
        \put(-1,5){\rotatebox{90}{\textbf{\texttt{GT}}}}
        \put(-1,15){\rotatebox{90}{\textbf{\texttt{Ours}}}}
        
        \put(-1,24.5){\rotatebox{90}{\textcolor[rgb]{0.99,0.5,0.0}{\textbf{\texttt{XMF}}}\cite{aiello2022cross}}}
        
        \put(-1,33){\rotatebox{90}{\textcolor[rgb]{0.99,0.5,0.0}{\textbf{\texttt{PFusion}}}\cite{xu2018pointfusion}}}
        
        \put(-1,46){\rotatebox{90}{\textcolor[rgb]{0.99,0.5,0.0}{\textbf{\texttt{ILN}}}\cite{chen2022imlovenet}}}
        
        \put(-1,55.5){\rotatebox{90}{\textcolor[rgb]{0.2,0.8,0.1}{\textbf{\texttt{FRCNN}}}\cite{xu2022fusionrcnn}}}

        \put(-1,67){\rotatebox{90}{\textcolor[rgb]{0.2,0.8,0.1}{\textbf{\texttt{PMF}}}\cite{zhuang2021perception}}}

        \put(-1,77.5){\rotatebox{90}{\textcolor[rgb]{0.2,0.8,0.1}{\textbf{\texttt{MBDF}}}\cite{tan2021mbdf}}}
	\end{overpic}
	\caption{\textbf{Visual results.} We give some visual results of all comparison methods and our method in the \textbf{\texttt{Unseen}}.}
	\label{comparative_unseen}
\end{figure*}

\begin{figure*}[t]
	\centering
		\begin{overpic}[width=0.86\linewidth]{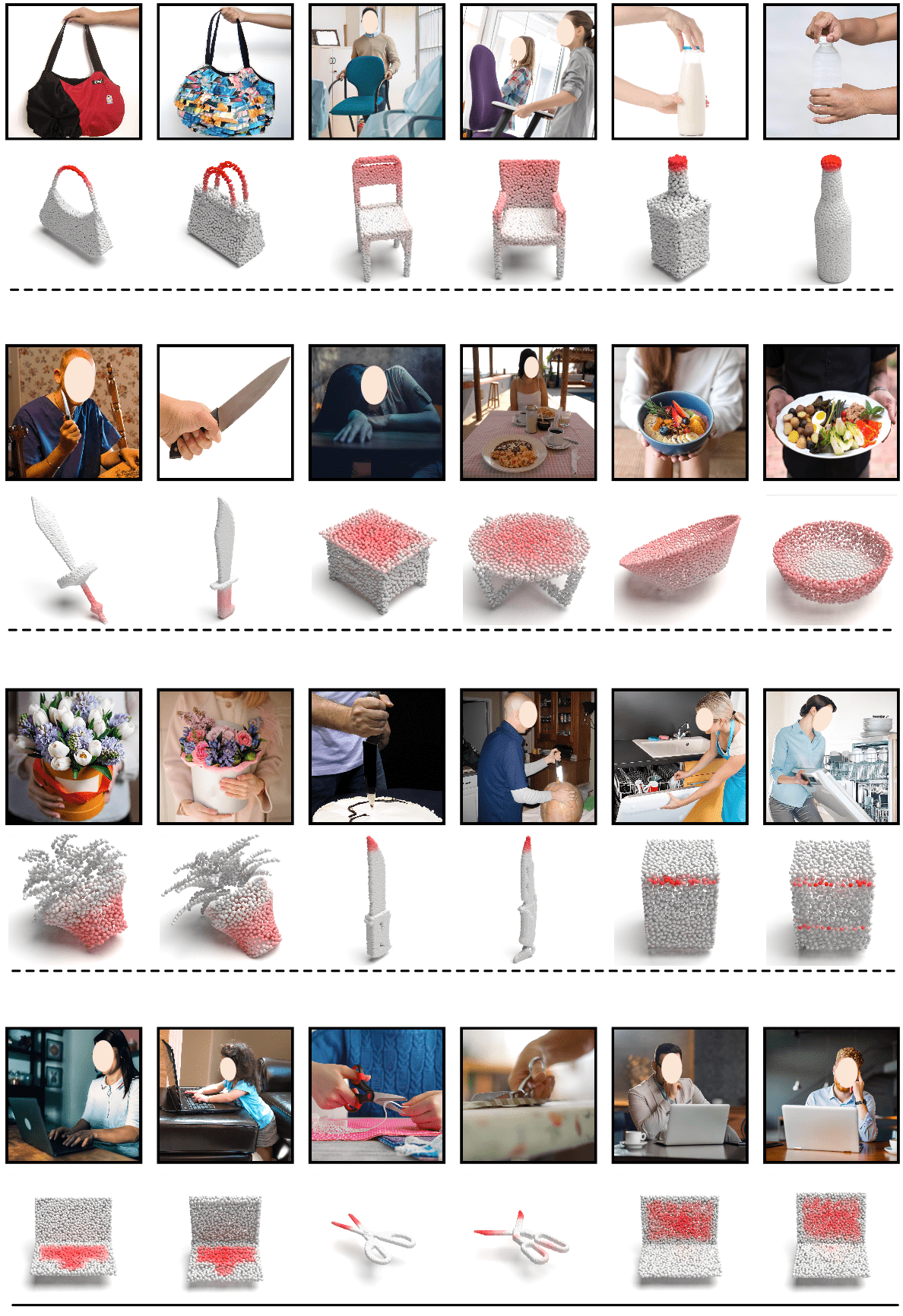}
            \put(4,100.5){\textbf{Lift}}
            \put(26.5,100.5){\textbf{Move}}
            \put(49.5,100.5){\textbf{Open}}

            \put(3,75){\textbf{Grasp}}
            \put(25.5,75){\textbf{Support}}
            \put(47,75){\textbf{Wrapgrasp}}

            \put(1.5,49){\textbf{Wrapgrasp}}
            \put(27,49){\textbf{Stab}}
            \put(49.5,49){\textbf{Open}}
            
            \put(3.5,23.5){\textbf{Press}}
            \put(27,23.5){\textbf{Cut}}
            \put(48.8,23.5){\textbf{Display}}
	\end{overpic}
	\caption{\textbf{Visual results of Our Method.} We give more results of our method in \textbf{\texttt{Seen}}, \textbf{\texttt{Unseen}}.}
	\label{Ours_results}
\end{figure*}

\begin{figure*}[t]
	\centering
		\begin{overpic}[width=0.75\linewidth]{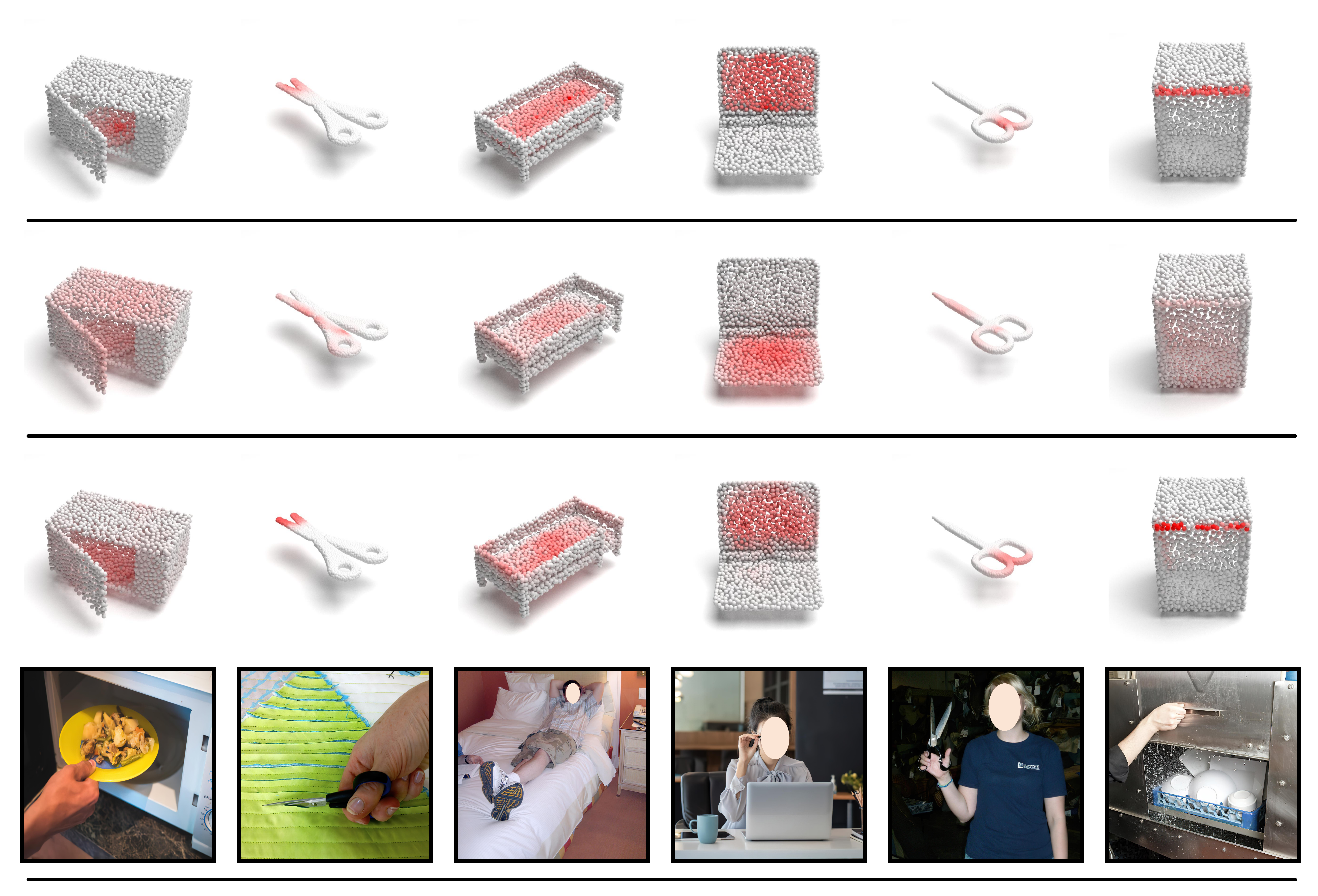}
            \put(4,66){\textbf{Contain}}
            \put(22.5,66){\textbf{Cut}}
            \put(38,66){\textbf{Lay}}
            \put(54,66){\textbf{Display}}
            \put(71,66){\textbf{Grasp}}
           \put(88,66){\textbf{Open}}
            \put(-1.5,57){\rotatebox{90}{\textbf{GT}}}
            \put(-1.5,39){\rotatebox{90}{\textbf{3DA.N.}}}
            \put(-1.5,15){\rotatebox{90}{\textbf{Ours}}}
	\end{overpic}
	\caption{\textbf{Generalization in \texttt{Unseen}}. 3D.A.N is 3D-AffordanceNet \cite{deng20213d}. Visualization results of this method and ours in \textbf{\texttt{Unseen}}.}
	\label{fig:paradigm}
\end{figure*}

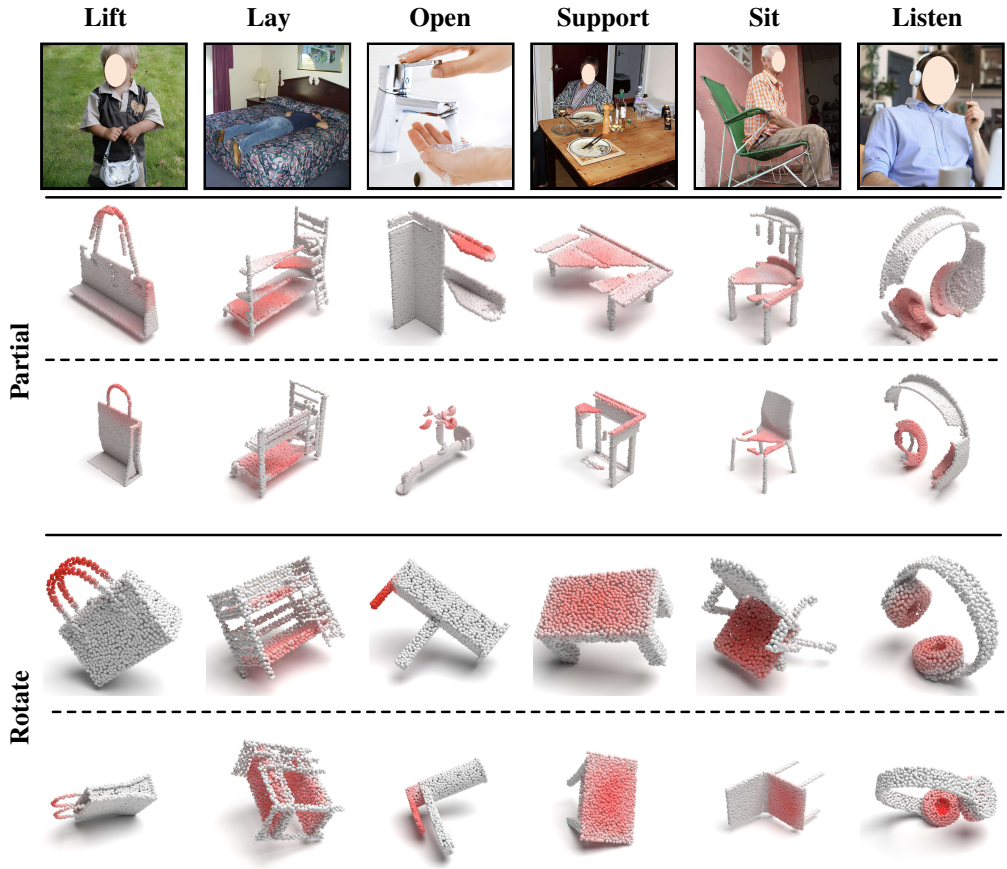
\begin{figure*}[t]
	\centering
		\begin{overpic}[width=0.75\linewidth]{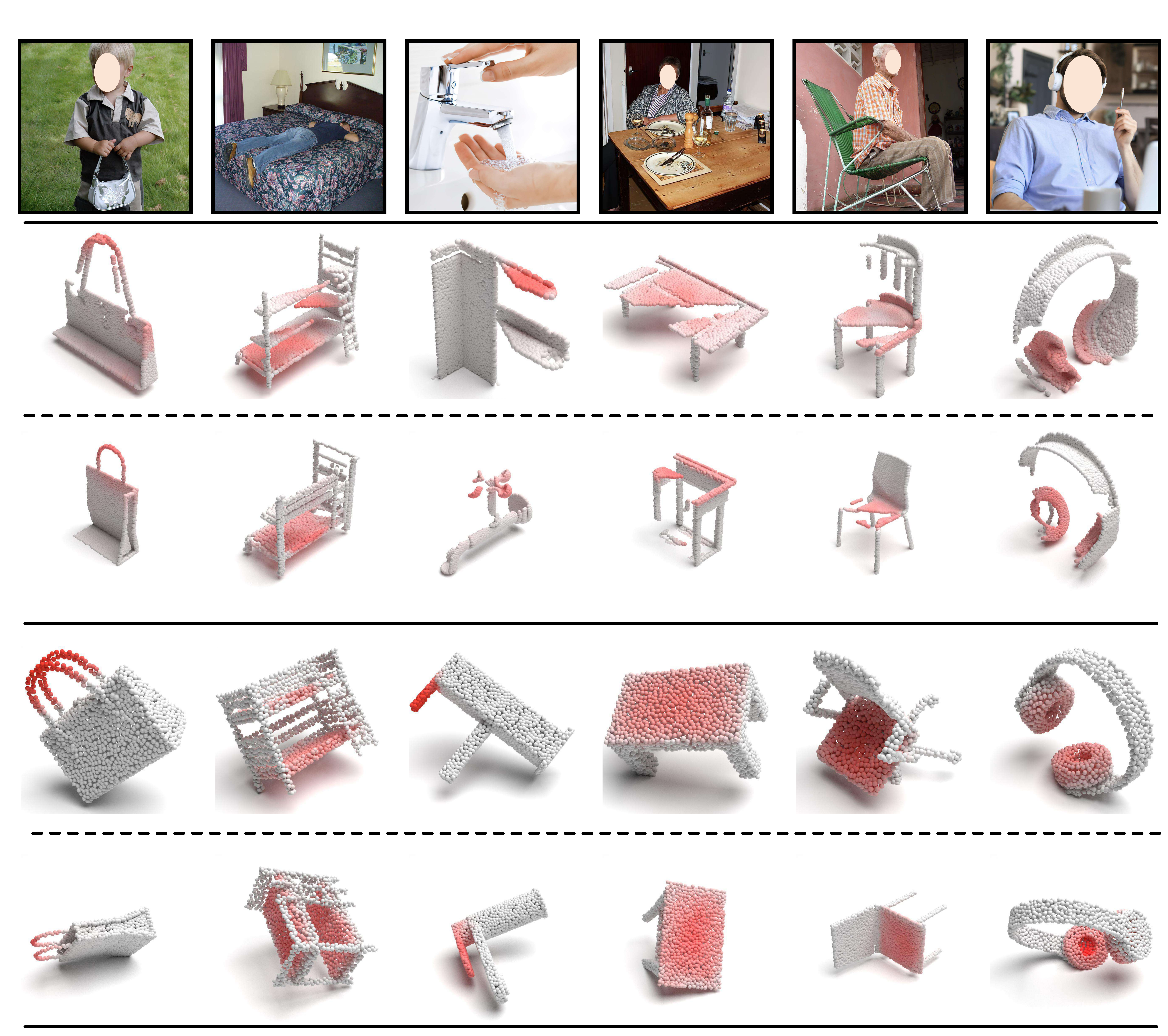}
            \put(6,87){\textbf{Lift}}
            \put(22.5,87){\textbf{Lay}}
            \put(39,87){\textbf{Open}}
            \put(54,87){\textbf{Support}}
            \put(73.5,87){\textbf{Sit}}
            \put(88,87){\textbf{Listen}}
            \put(-1.5,14){\rotatebox{90}{\textbf{Rotate}}}
            \put(-1.5,49){\rotatebox{90}{\textbf{Partial}}}
	\end{overpic}
	\caption{\textbf{Partial and Rotate.} Visualization results of partial point cloud and rotated point cloud.}
	\label{fig:rotate}
\end{figure*}

\end{document}